%% file: paper.tex
\documentclass{bmvc2k}

\title{Shape Preserving Facial Landmarks with Graph Attention Networks}

\usepackage{multirow}
\usepackage{tablefootnote}
\usepackage{amsfonts}

\addauthor{Andr\'es Prados-Torreblanca}{a.prados@upm.es}{1,2}
\addauthor{Jos\'e M. Buenaposada}{josemiguel.buenaposada@urjc.es}{1}
\addauthor{Luis Baumela}{lbaumela@fi.upm.es}{2}

\addinstitution{
 ETSII\\
 Universidad Rey Juan Carlos\\
 M\'ostoles, Spain
}
\addinstitution{
 Departamento de Inteligencia Artificial.\\
 Universidad Polit\'ecnica de Madrid,\\
 Boadilla del Monte, Spain
}

\runninghead{Prados, Buenaposada, Baumela}{Shape preserving facial landmarks}

\input{labvision_defs}

\newcommand{\first}[1]{{\color{blue} \textbf{#1}}}
\newcommand{\second}[1]{{\color{green} #1}}
\newcommand{\third}[1]{{\color{red} #1}}

\begin{document}

\maketitle

\begin{abstract}
Top-performing landmark estimation algorithms are based on exploiting the excellent ability of large convolutional neural networks (CNNs) to represent local appearance. However, it is well known that they can only learn weak spatial relationships. To address this problem, we propose a model based on the combination of a CNN with a cascade of Graph Attention Network regressors. To this end, we introduce an encoding that jointly represents the appearance and location of facial landmarks and an attention mechanism to weigh the information according to its reliability. This is combined with a multi-task approach to initialize the location of graph nodes and a coarse-to-fine landmark description scheme. Our experiments confirm that the proposed model learns a global representation of the structure of the face, achieving top performance in popular benchmarks on head pose and landmark estimation. The improvement provided by our model is most significant in situations involving large changes in the local appearance of landmarks. The code is publicly available at \href{https://github.com/andresprados/SPIGA}{\color{bmv@sectioncolor}{https://github.com/andresprados/SPIGA}}

\end{abstract}

\section{Introduction}
\label{sec:intro}

Landmarks (or keypoints) are a widely used representation to address high-level vision tasks such as image retrieval~\cite{MoskvyakWACV21}, facial expression recognition~\cite{Sun19}, face reenactment~\cite{Zhang20Freenet}, etc.
The performance of computer vision algorithms on the final task depends, to a great extent, on the accuracy and robustness of this intermediate representation. Thus, although many algorithms with excellent performance have recently emerged, research is still very intense in this area.

Top facial landmark estimation methods may be broadly grouped into coordinate and heatmap regression approaches. \textit{Coordinate regression approaches} directly estimate the landmark position by projecting the representation estimated by a CNN encoder onto a set of 2D coordinates~\cite{Feng18wing,Kowalski17,Feng20rwing,Trigeorgis16,LinTIP21}. They are the most efficient since they only require an encoder architecture to compute the facial representation. 
The \textit{heatmap regression approach} is based on appending multiple encoder-decoder modules to estimate a 2D data structure modeling the landmark position likelihood, the heatmap~\cite{Honari16,Wu18lab,Wang19Awing,Huang20propnet,Kumar20luvli,Huang21ADnet}. The landmark coordinates are typically estimated at the maximum of each heatmap. This architecture provides an increase in accuracy at the expense of a considerable boost in computational and memory requirements. A fundamental limitation of both approaches is their degradation when there is ambiguity or noise contaminating the local landmark appearance. This typically happens at the presence of occlusions, heavy make-up, blur and extreme illuminations or poses. This is because of the known fact that CNNs cannot learn simple spatial relationships~\cite{Santoro17} and, in the case of facial landmarks, are unable to learn a global representation of the face structure. However, a human face is a highly structured object with a prominent landmark configuration. Therefore, an effective way of representing the local appearance of each landmark and its geometric relationship to the other landmarks is needed.

This problem has been partially addressed in the literature with a local attention module combining landmarks with facial boundaries~\cite{Wu18lab,Huang20propnet,Huang21ADnet}. This is a solution that learns short-distance geometrical relationships. An alternative solution combines the advantages of a CNN description with traditional Ensemble of Regression Trees (ERT)~\cite{Valle18,Valle193dde}. Although this solution is able to learn long-distance geometrical dependencies, it is not fully satisfactory because of the limited learning capabilities of ERTs and the impossibility of end-to-end training.
Other approaches use a Graph Convolutional Network (GCN) to learn the facial geometrical structure~\cite{Li20sld,LinTIP21}. This is achieved by combining the landmark local description, extracted from the CNN representation, with geometrical information represented by the relative landmark locations. However, poor initialization and the lack of an advanced attention mechanism reduce the performance of these models.
More recent approaches use transformers~\cite{Li22casctransf, Xia22slpt} in a cascade shape regressor, obtaining very good results due to the built-in attention mechanisms.

In this paper, we present the SPIGA (\emph{Shape Preserving wIth GAts}) model for the estimation of human face landmarks. We follow the traditional regressor cascade approach~\cite{Cao14} and present an algorithm that combines a multi-stage heatmap backbone with a cascade of Graph Attention Network (GAT) regressors~\cite{Velickovic18gats}. The backbone provides a top-performing facial appearance representation. The cascaded GAT regressor is endowed with a positional encoding and attention mechanism that learn the geometrical relationship among landmarks. Another element of our proposal that improves the convergence of the GAT cascade is a coarse-to-fine feature extraction procedure and a good initialization. To do this, we train our backbone with a multi-task approach that also estimates the head pose, using its projection to establish the initial landmark locations. We evaluate the performance of our proposal in 300W, COFW-68, MERL-RAV and WFLW datasets. It achieves top performance on both head pose and face landmarks estimation. The improvement is most significant in situations involving large appearance changes, such as occlusions, heavy make-up, blur and extreme illuminations. 
We make the following contributions:
1) A GAT cascade with an attention mechanism to weigh the information provided by each landmark according to its reliability;
2) A positional encoding to jointly represent relative landmark locations and local appearance;
3) A multi-task approach to initialize the location of graph nodes;
4) A coarse-to-fine landmark description scheme.

\section{Shape Regressor Model}
\label{sec:our_method}

We propose a coarse-to-fine cascade of landmark regressors~\cite{Dollar10,Cao14} that iteratively refines the landmarks coordinates while preserving the face shape. Our approach involves three critical components: 1) the initialization, 2) the features used for regression, and 3) the regressors that estimate the face shape deformation at each step of the cascade. 

\begin{figure}
    \centering
    \includegraphics[width=\textwidth]{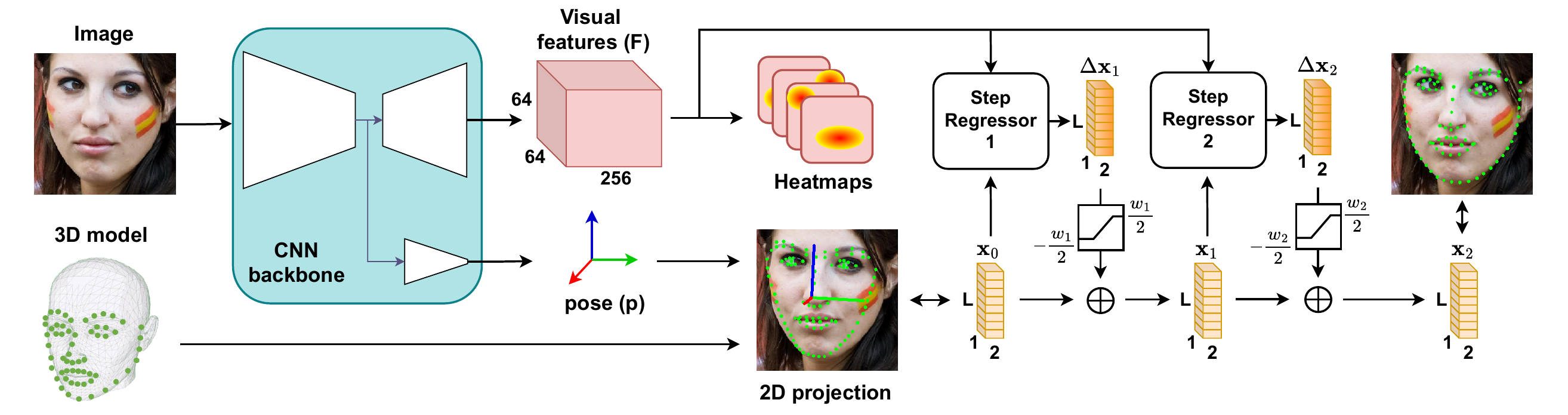}
    \caption{Regressor architecture with a two-step cascade.}
    \label{fig:cascade_regressor_full}
\end{figure}

In our proposal, we use a multi-task CNN backbone to provide both, the initialization and the local appearance representation. We set the initial shape of the face, $\vx_0\in\Reales^{L\times 2}$,  by projecting $L$ landmarks from a generic 3D rigid face mesh oriented using the head pose backbone prediction. At each cascade step $t$, a GAT-based~\cite{Velickovic18gats} regressor computes a displacement vector, $\Delta\vx_t$, to update the landmarks location, $\vx_t = \vx_{t-1} + \Delta\vx_t $. After $K$ steps, the final face shape is $\vx_K = \vx_0 + \sum_{t=1}^K \Delta\vx_t$. We denote the 2D location of $l$-th landmark at step $t$ as $\vx_t^l\in\Reales^{2}$. In Fig.~\ref{fig:cascade_regressor_full} we show the regressor with a two-step cascade configuration.

\subsection{Initialization by Head Pose Estimation}
\label{sec:init_3dpose}

Our multi-task backbone, termed \emph{Multi Task Network} (\emph{MTN}), is a cascade of $M$ encoder-decoder Hourglass (HG) 
modules.
Each HG module in MTN is composed of a shared encoder with two task branches: 
1) a 3D head pose estimation branch and 2) a landmark estimation decoder to the end of which we attach the next HG module. Defining and balancing the depth of the three components is a critical factor to boost the head pose estimation accuracy. We supervise the $h$-th module pose 
head by comparing its estimation, $\vp\in\Reales^6$, with the ground truth, $\tilde{\vp}$, using the L2 loss, $\cL_{\vp}^h(\vp, \tilde{\vp}) = ||\tilde{\vp} - \vp||^2$. Our annotations for pose, $\tilde{\vp}$, are obtained from the ground truth landmarks using a rigid head model (see Fig.~\ref{fig:cascade_regressor_full}). In the landmarks task we optimize a coordinate smooth L1 loss ($\cL_{coord}$) enhanced by a local attention mechanism ($\cL_{att}$) on the heatmaps, like~\cite{Wang19Awing,Huang21ADnet}. The final landmark loss is defined as $\cL_{lnd}=\sum_{h=1}^M 2^{h-1} (\lambda_{c}\cL_{coord}^h + \lambda_{att}\cL_{att}^h)$, where $\lambda$'s are scalars empirically optimized. For further details, please see the supplementary material.

To obtain a top-performing head pose estimation model (see Table~\ref{tab:pose})
we pre-train the network only with the landmark task, $\cL_{lnd}$, and fine-tune with both tasks, landmarks and pose, like~\cite{Valle21}. For multi-task fine-tuning we use the loss  $\cL_{mt} = \cL_{lnd} + \lambda_{\vp}\sum_{h=1}^M 2^{h-1}\cL_{\vp}^h$, where $\lambda_{\vp}$ is a hyperparameter. Although we use intermediate supervision at every HG module, the prediction of $\vp$ to estimate $\vx_0$, as well as the visual features, are extracted from the last module. Let $\vX\in\Reales^{L\times 3}$ be the 3D coordinates on the 3D head model that correspond to the $L$ 2D landmarks. If the pose estimated by the backbone is given by $\vp$, then the \emph{initial shape}, $\vx_0$, is computed by projecting the 3D model, $\vx_0 = \pi(\vX; \vp)$, where $\pi(\cdot)$ is the 3D$\rightarrow$2D projection function.

\subsection{Geometric and Visual Feature Extraction}
\label{sec:features}

For each step in the cascaded regressor, the input features are a combination of local appearance at each landmark (i.e. visual features) and global representation of the facial structure (i.e. geometric features). 
How visual and positional information is extracted and combined has a direct impact on the performance of the regressor (see Table \ref{tab:ablation_wflw_npe90}). 

\begin{figure}
    \centering
    \includegraphics[width=\textwidth]{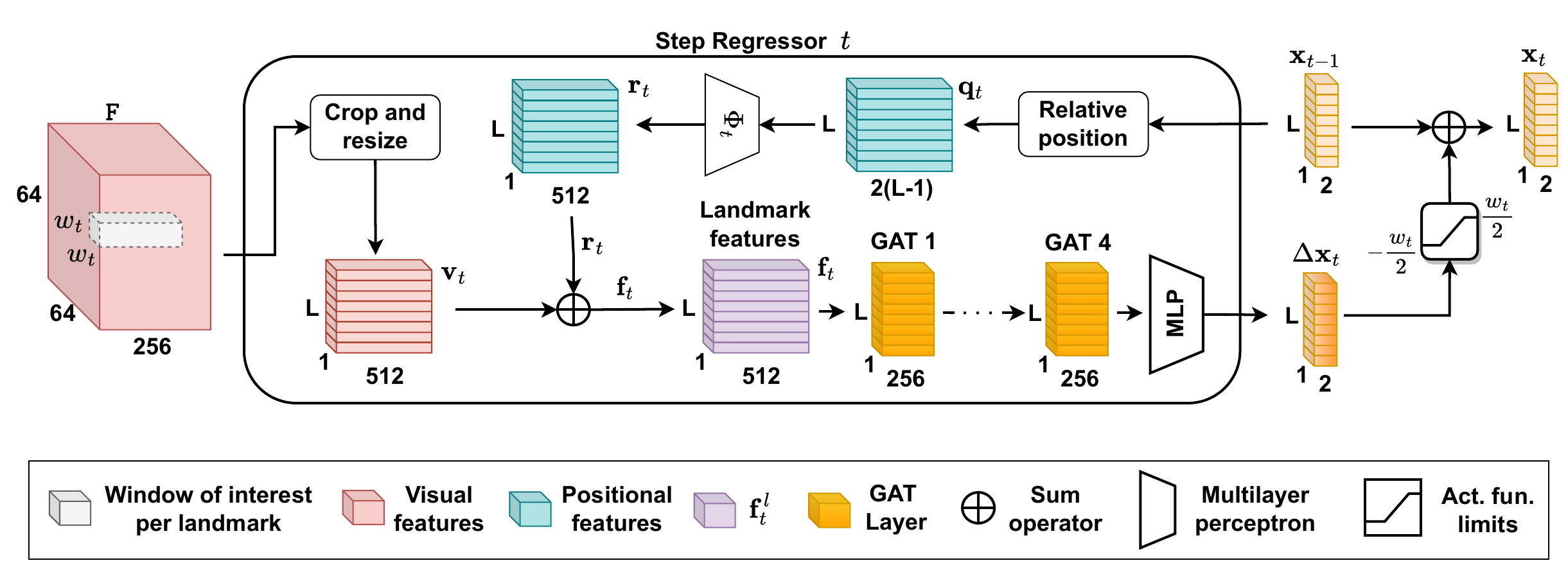}   
    \caption{Appearance and shape feature extraction for the $t$-th step regressor.}
    \label{fig:cascade_regressor_step}
\end{figure}

Let $\mF$ be the output feature map of the last stacked HG module in the MTN. We extract local appearance information from a square window, $\cW_t$, of size $w_t\times w_t$, centered at each landmark location, $\vx_{t-1}^l$, in $\mF$. We use a fixed affine transform with a grid generator and sampler~\cite{Jaderberg15spatial} to crop and re-sample $\cW_t$ at a fixed size, regardless of $w_t$. Then, using convolutional layers, we extract the visual features, $\vv_t^l$, corresponding to the $l$-th landmark at step $t$. We iteratively reduce $w_t$ at each step $t$, in a coarse-to-fine approach. 

Positional information is crucial to maintain the shape of the face when local appearance alone is not sufficient (e.g. in presence of occlusions, blur, make-up, etc.). Relative distances between landmarks provide enhanced geometrical features compared to their absolute locations since they explicitly represent the facial shape. This relative positional information can be defined from displacement vectors between landmarks~\cite{Li20sld}. Let $\vq_t^l=\{\vx_{t-1}^l - \vx_{t-1}^i \}_{i\neq l} \in \Reales^{2\times (L-1)}$ be the displacement vector corresponding to $l$-th landmark in the $t$-th step. In contrast to~\cite{Li20sld}, we learn a high dimensional embedding from $\vq_t^l$ using a Multi layer Perceptron (MLP), $\vr_t^l = \Phi_t(\vq_t^l)$, that facilitates the aggregation of the visual local appearance and the facial shape information. In the experiments, we show that this way of encoding relative positional information in $\vr$ improves the shape-preserving ability of the network (see section \ref{sec:ablation}).

Let $\vf_t^l$ be the feature vector used to compute $\Delta\vx_t^l$.
At each step $t$ of the cascade (see Fig.~\ref{fig:cascade_regressor_step}), and for each landmark $l$, we add the visual features extracted from the backbone network, $\vv_t^l$, with the relative positional features, $\vr_t^l$, computed from the current shape, $\vx_{t-1}$, to produce the encoded features, $\vf_t^l=\vv_t^l + \vr_t^l$.

\subsection{Cascade Shape Regressor Using GATs}
\label{sec:gnn_regressor}
The step regressor architecture (Fig.~\ref{fig:cascade_regressor_step}) is composed of stacked GAT layers inspired by the ones in the Attentional Graph Neural Net~\cite{Sarlin20superglue}. We consider the facial shape as a single densely connected graph where nodes are the landmark locations, $\vx_t$. To weigh the shared information across nodes, we compute a dynamic adjacency matrix per GAT layer $s$, $\cA_t^{s}$.  We learn these matrices as an attention from a given landmark to every other in the graph.

The input to the first GAT layer at step $t$ are the encoded features, $\{\vf_t^i\}_{i=1}^L$. Let $\vf_t^{i,s-1}$ be the features of the $i$-th landmark produced by the ($s$-1)-th GAT layer, that are also the input to $s$-th layer ($\vf_t^{i,0}\equiv\vf_t^i$). From now on, we drop the step-index $t$ to simplify the notation.
The updated feature vector after the $s$-th layer is defined as $\vf^{i,s} = \vf^{i,s-1} + MLP([\vf^{i,s-1} || \vm^{i,s}])$
where $[\cdot || \cdot]$ is the concatenation operator, $\vm^{i,s}$ is the information aggregated, or message, of the nodes neighboring $i$. Focusing on the message generation procedure, a query vector $\vh_q^{i,s}$, is assigned to landmark $i$ and key $\vh_k^{j,s}$, and value vectors $\vh_v^{j,s}$, to every other landmark $j$. The attention weight of landmark $i$ to landmark $j$ is the \texttt{SoftMax} over the key-query similarities $\alpha_{ij}=$ SoftMax$_j(\vh_q^{i,s} \cdot \vh_k^{j,s})$, being $\alpha_{ij}$ the elements of the adjacency matrix $\cA^s_t$ and the transmitted message $\vm^{i,s}$ the weighted average of the value vectors:
$\vm^{i,s} = \sum_{i \neq j} \alpha_{ij} \vh_v^{j,s}$,
where
$\vh_q^{i,s}=\mW_1^{s}\vf^{i,s} + \vb_1^s$, 
$\vh_k^{j,s}=\mW_2^{s}\vf^{j,s} + \vb_2^s$ and 
$\vh_v^{j,s}=\mW_3^{s}\vf^{j,s} + \vb_3^s$. 
Matrices $\mW_i$ and bias vectors $\vb_i$ are learned.

Finally, the last GAT layer output $\vf_t^{i,4}$ is processed by a decoder, an MLP, to obtain the corresponding displacement, $\Delta\vx_t^i$. We constraint the values in $\Delta\vx_t^i$, applying an \texttt{ArcTan} activation and scaling the result, to be in the interval $[-w_t/2, w_t/2]$. In practice, this constraint makes the single-step regressor search problem simpler, boosting training convergence. Given a trained MTN backbone, we train the cascade with the $\cL_{CR}=\sum_{t=1}^K L1_{smooth} [\tilde{\vx} - (\vx_{t-1} + \Delta\vx_t)]$ loss, where $\tilde{\vx}$ are the ground truth landmark coordinates.

\section{Experiments}
\label{sec:experiments}

To train and evaluate our method, we conduct different experiments in four complementary datasets which have been acquired in-the-wild and bear different levels of difficulty:

\textbf{300W}~\cite{Sagonas16} provides 68 manually annotated landmarks.
We employ the 300W private extension, which uses 3837 images as training set and adds 600 test images divided into indoor and outdoor subgroups.

\textbf{COFW-68} is a re-annotated version of COFW~\cite{Burgos13} with 68 landmarks. It is conceived for testing landmark detectors with occlusions in a cross-dataset approach. The testing set in COFW-68 is made of 507 images. The annotations include the landmark positions and the visibility labels for the same 68 points as in 300W.

\textbf{WFLW}~\cite{Wu18lab} is composed of challenging in-the-wild images and provides 98 manually annotated landmarks. The dataset has 7500 training and 2500 testing faces. It is divided into 6 subgroups: pose, expression, illumination, make-up, occlusion and blur.

\textbf{MERL-RAV}~\cite{Kumar20luvli} is a re-annotated version of 19,000 AFLW images with 68 landmarks, like 300W. It provides 15,449 training and 3,865 test faces divided into 3 orientation subsets: frontal, half-profile and profile. This recent dataset includes externally occluded visibility and self-occluded labels.

\subsection{Evaluation Metrics}
In order to quantify the head pose estimation error, we use the Mean Absolute Error (MAE) metric,
$MAE = \frac{1}{N} \sum\limits_{i=1}^{N}|\tilde{p}_{i} - p_{i}|$, 
where $N$ is the number of testing images, $\tilde{p}_{i}$ is the ground truth and $p_{i}$ represents a single predicted pose parameter. 

Focusing on the landmark estimation task, Normalized Mean Error (NME) is the standard metric, 
$NME = \frac{100}{N} \sum\limits_{i=1}^{N}\sum\limits_{l=1}^{L}\frac{||\tilde{\vx}_{l}^{i} - \vx_{l}^{i} ||_2}{d_{i}}$.
Where $\tilde{\vx}_{l}^{i}$ and $\vx_{l}^{i}$ denote, respectively, the ground-truth and predicted coordinates of the $i$-th landmark and $d_{i}$ is a normalization value which varies depending on the dataset: inter-ocular (int-ocul), distance between outer eye corners; inter-pupils, distance between pupil/eye centers; and box, computed as the geometric mean of the landmarks ground truth bounding box ($d=\sqrt{w_{bbox}*h_{bbox}}$). 

We also use Failure Rate (FR) and Area Under the Curve (AUC). FR evaluates the robustness of algorithms in terms of NME, indicating the percentage of images with an NME above a given threshold. AUC is calculated by computing the area under the Cumulative Error Distribution (CED) curve from 0 to the FR threshold. We introduce the Normalized mean Percentile Error 90 ($NPE_{90}$) which represents the NME for the image at the 90\% of the dataset, sorted by NME. This metric is particularly convenient for small data subsets where the FR is not representative. 

In all our tables results ranked \first{first}, \second{second} and \third{third} are shown respectively in blue, green and red colors.

\subsection{3D Pose Estimation Results}

First, we evaluate the MTN performance in 3D pose estimation. In Table~\ref{tab:pose}, we compare our pose estimation in 300W and WFLW with previous works in the literature. Our model shows a significant improvement. We reduce the mean MAE of the previous top performer, MNN~\cite{Valle21}, by 17\% and 27\% respectively in 300W and WFLW. The main reason behind this improvement is a better network architecture, stacked HGs vs. a single encoder-decoder in~\cite{Valle21} and the use of an attention mechanism.
Having such a precise head pose estimation is a critical factor in our proposal, since the cascade shape regressor initialization relies on this prediction. 

\begin{table}
\footnotesize
\begin{center}
\setlength{\tabcolsep}{4pt} 
\begin{tabular}{l|ccc|c|ccc|c|ccc|c}
\hline
       & \multicolumn{4}{c}{300W}   & \multicolumn{4}{c}{WFLW} & \multicolumn{4}{c}{MERL-RAV} \\
       & \multicolumn{4}{c}{Angular error $(^\circ)(\downarrow)$}   & \multicolumn{4}{c}{Angular error $(^\circ)(\downarrow)$} & \multicolumn{4}{c}{Angular error $(^\circ)(\downarrow)$} \\
Method & yaw & pitch & roll & mean  & yaw & pitch & roll & mean & yaw & pitch & roll & mean \\
\hline
Yang~\cite{Yang15}         & 4.2 & 5.1 & 2.4 & 3.9 & - & - & - & - & - & - & - & - \\
JFA~\cite{Xu17JFA}         & 2.5 & 3.0 & 2.6 & 2.7 & - & - & - & - & - & - & - & - \\
ASMNet~\cite{Fard21AMSNet} & 1.62  & 1.80 & 1.24 & 1.55 & 2.97  & 2.93 & 2.21 & 2.70 & - & - & - & - \\
MNN~\cite{Valle21}         &  -  & - &  - &  1.56 & - & - & - & 2.08 & - & - & - & - \\
\hline
\textbf{SPIGA (Ours)}      &  \first{1.41}  & \first{1.70} &  \first{0.77} &  \first{1.29}
                           & \first{1.78} & \first{1.86} & \first{0.93} & \first{1.52} 
                           & \first{3.23} & \first{2.24} & \first{1.71} & \first{2.39}\\
\hline
\end{tabular}
\end{center}
\caption{Head pose MAE, in degrees, for 300W public, WFLW and MERL-RAV datasets.}
\label{tab:pose}
\end{table}

\subsection{Landmark Detection Results}
\label{landmars_sota}

WFLW is the most popular benchmark to evaluate the performance of facial landmark detection. Recent methods that adopt this dataset use the bounding boxes provided by HRnet~\cite{Wang21hrnet}, that were obtained from the ground truth landmark annotations.
By doing so, they achieve better performance (see  Table~\ref{tab:wflw}, AWing results improve from 4.36 to 4.21 NME). In Table~\ref{tab:wflw}, we clearly distinguish the bounding boxes used in the evaluation. Another important aspect to perform a fair comparison is the use of additional training data. 
In our discussion we do not consider methods that train with images or annotations other than those provided by WFLW. 

In Table~\ref{tab:wflw}, we show that our model outperforms current state-of-the-art (SOTA) in most of the WFLW subsets, as well as in the full set metrics. When it is compared with other GraphNets-based methods, our approach is 4\% and 32\% better in terms of NME and FR than  SLD~\cite{Li20sld}, and 7\% and 23\% better than SDFL~\cite{LinTIP21}. These results show that our relative positional encoding and the per layer graph attention mechanism have a strong impact on the performance of GraphNets. Further, our proposal is also more accurate than recent approaches based on transformers, when these models are trained only with WFLW data,  DTLD-s~\cite{Li22casctransf} and SPLT~\cite{Xia22slpt}, both with 4.14 NME in the full set. 
If we analyze the performance on some of the subsets, our method is 35\%, 25\%, 23\% and 39\% better than the previous SOTA, ADNet~\cite{Huang21ADnet}, in the illumination, make-up, occlusion and blur subsets. This proves the importance of learning a global representation of the facial structure, that CNNs alone do not provide.
Additionally, the low FR across the different subsets and better AUC values reaffirm that our model achieves a balanced trade-off between robustness and precision,  taking advantage of the complementary benefits from the CNN and GAT architectures.

\begin{table}
\scriptsize
\begin{center}
\setlength{\tabcolsep}{4pt} 
\begin{tabular}{c|c|ccccccc}
\hline 
Metric & Method & Testset & Pose & Expression & Illumination & Make-up & Occlusion & Blur \\
\hline
\hline 
&\multicolumn{8}{c}{Bounding boxes from WFLW benchmark} \\
\cline{2-9} 
\multirow{16}{*}{$NME_{int-ocul}$ (\%)($\downarrow$)}  
& 3DDE~\cite{Valle193dde}                   & 4.68 & 8.62  & 5.21 & 4.65 & 4.60 & 5.77 & 5.41 \\
& DeCaFA~\cite{Dapogny19decafa}             & 4.62 & 8.11  & 4.65 & 4.41 & 4.63 & 5.74 & 5.38 \\
& AVS+SAN~\cite{Qian19Avs}                  & 4.39 & 8.42  & 4.68 & 4.24 & 4.37 & 5.60 & 4.86 \\
& AWing~\cite{Wang19Awing}                  & 4.36 & 7.38  & 4.58 & 4.32 & 4.27 & 5.19 & 4.96 \\
\cline{2-9}
&\multicolumn{8}{c}{Bounding boxes from GT landmarks (HRnet~\cite{Wang21hrnet} annotations)} \\
\cline{2-9}
& GlomFace~\cite{Zhu22glomface}             & 4.81 & 8.17  & - & - & - & 5.14 & - \\
& LUVLI~\cite{Kumar20luvli}                 & 4.37 & 7.56  & 4.77 & 4.30 & 4.33 & 5.29 & 4.94\\
& SDFL~\cite{LinTIP21}                 & 4.35 & 7.42  & 4.63 & 4.29 & 4.22 & 5.19 & 5.08\\
& AWing~\cite{Wang19Awing}              & 4.21 & \third{7.21}  & 4.46 & 4.23 & 4.02 & \third{4.99} & 4.82 \\ 
& SLD~\cite{Li20sld}                        & 4.21 & 7.36  & 4.49 & 4.12 & 4.05 & \second{4.98} & 4.82 \\
& HIHc\tablefootnote{Use RetinaFace detections.}~\cite{Lan21hih} & \third{4.18} & 7.20  & \first{4.19} & 4.45 & \second{3.97} & 5.00 & \third{4.81} \\
& ADNet~\cite{Huang21ADnet}                 & \second{4.14} & \first{6.96} & \second{4.38} & \third{4.09} & 4.05 & 5.06 & \second{4.79} \\
& DTLD-s~\cite{Li22casctransf}             & \second{4.14} & - & - & - & - & - & -\\
& SPLT~\cite{Xia22slpt}                 & \second{4.14} & \first{6.96} & \third{4.45} & \second{4.05} & \third{4.00} & 5.06 & \second{4.79}\\
\cline{2-9}
& \textbf{SPIGA (Ours)}       & \first{4.06} & \second{7.14} & \textbf{4.46} & \first{4.00} & \first{3.81} & \first{4.95} & \first{4.65}\\
\hline
\hline 
\multirow{7}{*}{$FR_{10}$ (\%)($\downarrow$)} 
& GlomFace~\cite{Zhu22glomface}             & 3.77 & 17.48  & - & - & - & 6.73 & - \\
& DTLD-s~\cite{Li22casctransf}             & 3.44 & - & - & - & - & - & -\\
& LUVLI~\cite{Kumar20luvli}                 & 3.12 & 15.95 & 3.18 & 2.15 & 3.40 & 6.39 & \third{3.23} \\
& SDFL~\cite{LinTIP21}               & 2.72 & 12.88  & \second{1.59} & 2.58 & 2.43 & 5.71 & 3.62\\
& AWing~\cite{Wang19Awing}                  & \first{2.04} & \first{9.20} & \first{1.27} & \third{2.01} & \first{0.97} & \first{4.21} & \second{2.72} \\ 
& SLD~\cite{Li20sld}                        & 3.04 & 15.95 & 2.86 & 2.72 & \second{1.46} & \third{5.29} & 4.01 \\
& HIHc$^1$~\cite{Lan21hih}                      & 2.96 & 15.03 & \second{1.59} & 2.58 & \second{1.46} & 6.11 & 3.49 \\
& ADNet~\cite{Huang21ADnet}         & \third{2.72} & 12.72 & \third{2.15} & 2.44 & \third{1.94} & 5.79 & 3.54 \\
& SPLT~\cite{Xia22slpt}             & 2.76 & \third{12.27} & 2.23 & \second{1.86} & 3.40 & 5.98 & 3.88\\
\cline{2-9}
& \textbf{SPIGA (ours)}             & \second{2.08} & \second{11.66} & \textbf{2.23} & \first{1.58} & \second{1.46} & \second{4.48} & \first{2.20}\\
\hline
\hline 
\multirow{4}{*}{$AUC_{10}$ (\%)($\uparrow$)} 
& AWing~\cite{Wang19Awing}                  & 58.95 & 33.37 & 57.18 & 59.58 & 60.17 & \third{52.75} & 53.93 \\ 
& SLD~\cite{Li20sld}                        & 58.93 & 31.50 & 56.63 & 59.53 & 60.38 & 52.35 & 53.29 \\
& HIHc$^1$~\cite{Lan21hih}                      & \third{59.70} & 34.20 & \first{59.00} & \second{60.60} & \third{60.40} & 52.70 & \second{54.90} \\
& ADNet~\cite{Huang21ADnet}                 & \second{60.22} & \third{34.41} & 52.34 & 58.05 & 60.07 & \second{52.95} & \third{54.80} \\
& SPLT~\cite{Xia22slpt}                 & 59.50 & \second{34.80} & \third{57.40} & \third{60.10} & \second{60.50} & 51.50 & 53.50\\
\cline{2-9}
& \textbf{SPIGA (Ours)}                             & \first{60.56} & \first{35.31} & \second{57.97} & \first{61.31} & \first{62.24} & \first{53.31} & \first{55.31}\\
\hline
\end{tabular}
\end{center}
\caption{Evaluation of landmark detection on WFLW.}
\label{tab:wflw}
\end{table}

\begin{figure}
    \centering
    \includegraphics[width=0.15\textwidth]{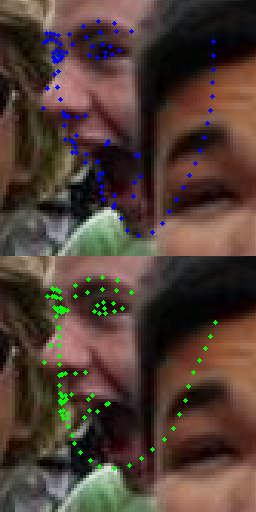}
    \includegraphics[width=0.15\textwidth]{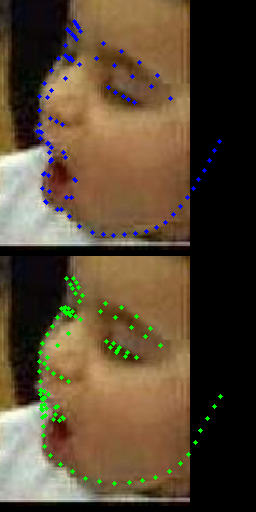}
    \includegraphics[width=0.15\textwidth]{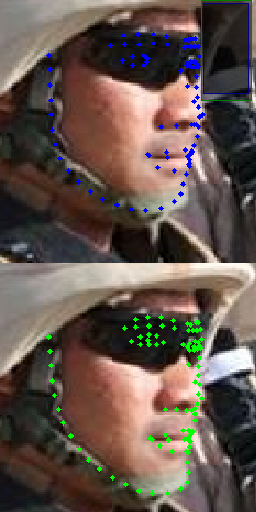}
    \includegraphics[width=0.15\textwidth]{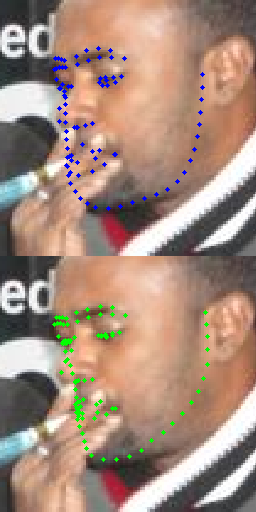}
    \includegraphics[width=0.15\textwidth]{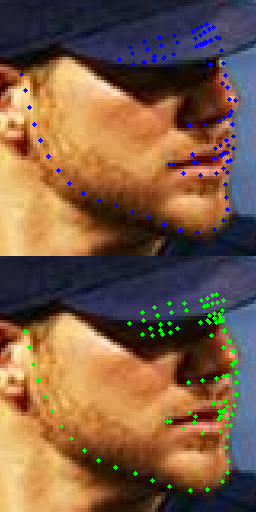}
    \includegraphics[width=0.15\textwidth]{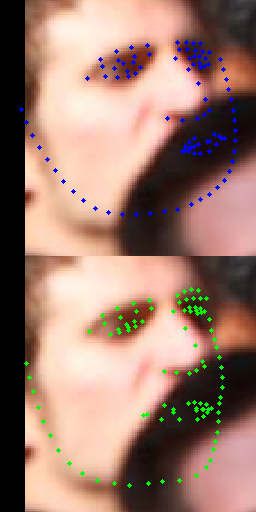}
    \caption{WFLW results on expressions (first 2 cols.) and pose examples (last 4 cols.). Shown in blue the ground truth and in green estimated landmarks.}
    \label{fig:expresion_pose_wflw_qualitative}
\end{figure}

On the other hand, results of subsets where our approach is not competitive also bear some relevant insights. First, further research is needed in the expression subset, where our performance is not as good as the rest. This is due to the fact that the 3D facial model used to initialize the cascade is rigid (see Fig.~\ref{fig:expresion_pose_wflw_qualitative}).
Second, seemingly, in the pose subset, we are not the top performers. However, as we can see in Fig.~\ref{fig:expresion_pose_wflw_qualitative}, faces with extreme poses are not well annotated and self-occlusions are not marked. So, the evaluation on this subset of WFLW is questionable.

MERL-RAV is one of the newest datasets, created to evaluate 2D facial alignment in-the-wild. It improves landmark annotations at half-profile and profile images by labeling the self-occlusion of landmarks.  
Hence, this dataset allows to correctly measure the performance of landmark detectors on samples with extreme poses. 
As we can see in Table~\ref{tab:merl_rav}, in terms of NME$_{box}$, our model is 6\% better than LUVLI's~\cite{Kumar20luvli} baseline, performing the best in all pose subsets. 

\begin{table}
\scriptsize
\begin{center}
\setlength{\tabcolsep}{4pt} 
\begin{tabular}{c|cccc|cccc}
  & \multicolumn{4}{c}{NME$_{box}$(\%)($\downarrow$)} & \multicolumn{4}{|c}{AUC$^{7}_{box}$(\%)($\uparrow$)} \\ 
  Method & All & Frontal & Half-Prof. & Profile & All & Frontal & Half-Prof.& Profile \\ 
\hline
DU-Net & 1.99 & 1.89 & 2.50 & 1.92 & 71.80 & 73.25 & 64.78 & 72.79\\ 
LUVLI~\cite{Kumar20luvli} & 1.61 & 1.74 & 1.79 & 1.25 & 77.08 & 75.33 & 74.69 & 82.10 \\
\hline
\textbf{SPIGA (Ours)} & \first{1.51} & \first{1.62} & \first{1.68} & \first{1.19} & \first{78.47} & \first{76.96} & \first{75.64} & \first{83.00} \\ 
\hline
\end{tabular}
\end{center}
\caption{Evaluation of landmark detection on MERL-RAV.}
\label{tab:merl_rav}
\end{table}

Finally, to verify the generalization and performance against occlusions, we conduct a cross-dataset experiment training with the 300W public split and testing with COFW-68 and 300W private. Results are summarized in Table~\ref{tab:cofw68_300w_private}. They prove the importance of the graph attention mechanism, which dynamically weighs landmark relationships according to the local image appearance and relative position, versus a learned static relationship approach, such as SLD~\cite{Li20sld}, ($NME_{int-ocul}$ of 3.93 vs 4.22 in COFW-68). Further, SPIGA trained on the 300W public dataset beats LUVLI~\cite{Kumar20luvli} ($NME_{box}$ of 2.52 vs 2.75 in COFW-68) with a backbone that has half the number of HG modules. 
It also obtains comparable results to a recent transformer-based method trained from scratch, DTLD-s~\cite{Li22casctransf}. It is marginally better than DTLD-s in 300W private and worse in COFW-68.
These results prove that a general architecture using GATs can complement and enhance CNN-based models, reaching better results in situations where ambiguity or noise is contaminating the local landmark appearance, where preserving structural landmarks consistency contributes to the final solution.

\begin{table}
\scriptsize
\setlength{\tabcolsep}{4pt} 
\begin{center}
\begin{tabular}{l|cc|cc|c}
\hline
       & \multicolumn{2}{c|}{$NME_{box}$ (\%)($\downarrow$)}  
       & \multicolumn{2}{c|}{$AUC_{box}^7$ (\%)($\uparrow$)} 
       & \multicolumn{1}{c}{$NME_{int-ocul}$ (\%)($\downarrow$)}  \\ 
       & 300W priv. & COFW-68 & 300W priv. & COFW-68  & COFW-68 \\ 
\hline
HRNetV2-W18~\cite{Wang21hrnet}    & -    & -    & -    & -    & 5.06  \\ 
HG$\times$1+SAAT \cite{Zhu21adversarial} & -    & -    & -    & -  & 4.61  \\ 
LUVLI(8)~\cite{Kumar20luvli}      & 2.24 & 2.75 & 68.3 & 60.8 & - \\ 
GlomFace~\cite{Zhu22glomface}     & -    & 2.69~\tablefootnote{Result comes from a personal communication with authors of~\cite{Zhu22glomface}, 2.09 mistakenly in the paper.} & -    & -    & 4.21 \\ 
SLD~\cite{Li20sld}                & -    & -    & -    & -    & 4.22  \\
SDFL~\cite{LinTIP21}              & -    & -    & -    & -   & 4.18  \\
SPLT~\cite{Xia22slpt}           & -    & -    & -    & -    & 4.10 \\
DTLD-s~\cite{Li22casctransf}      & 2.05 & \first{2.47} & 70.9 & \first{65.0} & - \\ 
\hline
\textbf{SPIGA(4) (ours)}               & \first{2.03} & 2.52 & \first{71.0} & 64.1 & \first{3.93} \\
\hline
\end{tabular}
\end{center}
\caption{Landmark detection results on 300W private and COFW-68. In (·) we show the number of HG modules.}
\label{tab:cofw68_300w_private}
\end{table}

\subsection{Ablation Study}
\label{sec:ablation}

\begin{figure}
    \centering
    \includegraphics[width=\columnwidth]{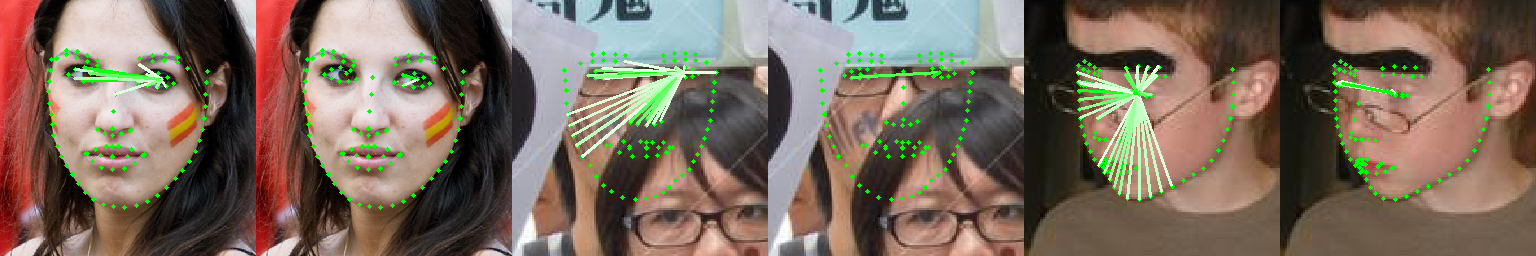}
    \caption{Left pupil attention mechanism at first and last layer, respectively, of the first regressor step.}
    \label{fig:attention}
\end{figure}

We conduct our ablation study on WFLW to understand how SPIGA components impact specific subset metrics. 
Table~\ref{tab:ablation_wflw_npe90} shows that the addition of the cascade shape regressor outperforms the bare MTN backbone (using \texttt{SoftArgMax}). 
Our new relative positional encoding is better than stacking the vector $\vq_t^l$ with the visual features,
and much better than using no positional information. The estimation of an attention per layer with the GAT improves with respect to use of a common attention matrix (GCN). An extended view of the effect of the learned adjacency matrix is shown in Fig.~\ref{fig:attention}. Occlusion images show how the attention mechanism relies on visible landmarks regardless of the layer. The regressor "looks" at distant and unoccluded landmarks at the first GAT layer and then at closer ones in the last layers. The contribution of the proposed coarse-to-fine scheme w.r.t. a constant size window ($w=8$) or a single pixel window ($w=1$) is also clear in Table~\ref{tab:ablation_wflw_npe90}. The improvement provided by SPIGA can be seen across all metrics. However, it is more prominent with the hard cases, as demonstrated by the results for the subsets  Makeup, Occlusion, and Blur, and the $NPE_{90}$ of the full set.

\begin{table*}
\scriptsize
\begin{center}
\setlength\tabcolsep{2pt}
\begin{tabular}{c|c|cc|cc|cc|cc}
\hline 
\multicolumn{2}{c|}{Changed from SPIGA model:} & \multicolumn{2}{c}{Full} & \multicolumn{2}{c}{Make-up}  & \multicolumn{2}{c}{Occlusion} & \multicolumn{2}{c}{Blur}  \\
Changed       & From $\rightarrow$ To   & $NME$ & $NPE_{90}$ & $NME$ & $NPE_{90}$ & $NME$ & $NPE_{90}$ & $NME$ & $NPE_{90}$ \\
\hline
Shape model & SPIGA $\rightarrow$ MTN backbone 
& 4.13 & 6.93  & 4.06 & 7.43 & 5.10 & 8.58 & 4.81 & 7.70  \\
\hline
\multirow{2}{*}{Positional encoding} & SPIGA $\rightarrow$ w/o pos. encod. & 4.17 & 7.07  & 4.01 & 6.71 & 5.03 & 8.33 & 4.72 & 7.52 \\
 & SPIGA $\rightarrow$ stacking & 4.09 & 6.87  & \third{3.83} & \second{6.47} & \second{4.97} & 8.15 & \third{4.68} & \second{7.37} \\
\hline
Attention & GAT $\rightarrow$ GCN & \second{4.08} & \second{6.79}  & 3.84 & 6.54 & \third{4.98} & \first{8.05} & \third{4.68} & \second{7.37} \\
\hline
\multirow{2}{*}{Coarse-to-Fine} & $w=16,8,4$ $\rightarrow$ $w=1,1,1$ & 4.12 & 6.95  & 3.88 & 6.76 & 4.99 & 8.19 & 4.71 & 7.44  \\
 & $w=16,8,4$ $\rightarrow$ $w=8,8,8$ & \second{4.08} & \third{6.84}  & \second{3.82} & \third{6.53} & \third{4.98} & \third{8.13} & \second{4.67} & 7.43  \\
\hline
- & \textbf{Best SPIGA model}  & \first{4.06} & \first{6.76}  & \first{3.81} & \first{6.32} & \first{4.95} & \second{8.09} & \first{4.65} & \first{7.31}  \\
\hline
\end{tabular}
\end{center}
\caption{
Contribution of the SPIGA components to the $NME_{int-ocul}$($\downarrow$) and $NPE_{90}$($\downarrow$) in WFLW.} 
\label{tab:ablation_wflw_npe90}
\end{table*}

In each row of Table~\ref{tab:ablation_step}, we display respectively the performance of three SPIGA models configured with one, two and three steps cascade. In each column, we show the NME obtained at each step. The final NME is reduced gradually as we increase the number of steps. Further, shorter cascades tend to have a better NME at the first step (4.17 vs 4.22). However, given also the larger FR they achieve (2.60 vs 2.44), we can conclude that longer cascades focus their first steps on improving their robustness.

\begin{table}
\scriptsize
\begin{center}
\begin{tabular}{c|ccc|ccc|ccc}
\hline 
 & \multicolumn{3}{c}{Step 1} & \multicolumn{3}{|c}{Step 2} & \multicolumn{3}{|c}{Step 3}  \\
Method & $NME_{int-ocul}$ & $AUC_{10}$ & $FR_{10}$ & $NME_{int-ocul}$ & $AUC_{10}$ & $FR_{10}$ & $NME_{int-ocul}$ & $AUC_{10}$ & $FR_{10}$ \\
& $(\downarrow)$ & $(\uparrow)$ & $(\downarrow)$ & $(\downarrow)$ & $(\uparrow)$ & $(\downarrow)$ & $(\downarrow)$ & $(\uparrow)$ & $(\downarrow)$ \\
\hline
SPIGA(1) & 4.17 & 59.53 & 2.60 & - & - & - & - & - & - \\
SPIGA(2) & 4.17 & 59.55 & 2.44 & 4.07 & 60.45 & 2.20 & - & - & - \\
SPIGA(3) & 4.22 & 59.10 & 2.44 & 4.08 & 60.41 & 2.12 & 4.06 & 60.56 & 2.08 \\
\hline
\end{tabular}
\end{center}
\caption{SPIGA results for cascades with different number of steps, shown in ().}
\label{tab:ablation_step}
\end{table}

\begin{figure}
    \centering
    \includegraphics[width=0.9\columnwidth]{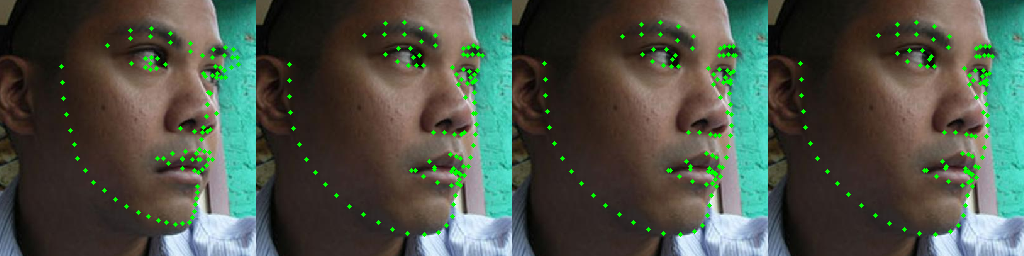}
    \includegraphics[width=0.9\columnwidth]{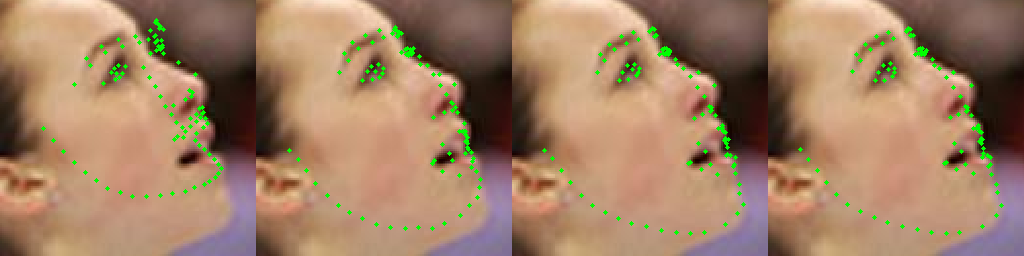}
    \caption{Estimated landmark locations: from 2D projection of the rigid 3D model (left) to the final result after the 3 regressor steps (right).}
    \label{fig:regressor_steps}
\end{figure}

In Fig.~\ref{fig:regressor_steps} we show the initialization and the landmark locations estimated at each step of the regressor cascade. When the face displays a neutral expression (top row), the initialization is reasonably good and the model converges to a solution within one regression step. Since SPIGA initializes landmarks with a 3D model featuring a neutral expression, when the face displays any other configuration, the initialization is much worse (lower row). However, even in this situation, the model is able to estimate the correct landmark locations in three regression steps. 

\section{Conclusions}
\label{sec:conclussions}

We presented SPIGA, a face landmark regressor that combines a CNN with a cascade of Graph Attention Networks (GATs). The CNN provides the local appearance representation. The GAT regressor is endowed with a positional encoding and attention mechanism that learn the geometrical relationship among landmarks and encourage the model to produce plausible face shapes. It establishes a new SOTA in the WLFW, COFW-68 and MERL-RAV datasets. In our experimentation we verify that the positional encoding is the component that contributes most to the final result and the first steps of the cascade focus on improving the robustness. 
In addition, at each step, the regressor "looks" at distant and reliable landmarks in the first GAT layer and progressively focuses its attention on closer landmarks in the following ones.  These insights from our ablation analysis confirm that SPIGA is learning a global representation and explains why its improvement is most significative in challenging situations involving occlusions, heavy make-up, blur and extreme illumination.

\section*{Acknowledgements}

The following funding is gratefully acknowledged. Andr{\'e}s Prados was funded by the Comunidad de Madrid, Ayudantes de Investigaci{\'o}n grant PEJ-2019-AI/TIC-15032. Jos{\'e} M. Buenaposada is funded by the Comunidad de Madrid project RoboCity2030-DIH-CM (S2018 /NMT-4331).

\bibliography{faces}
\end{document}


\maketitle

\section{Implementation Details}
\label{sec:implemtation_details}
\begin{figure}[h]
    \centering
    \includegraphics[width=\columnwidth]{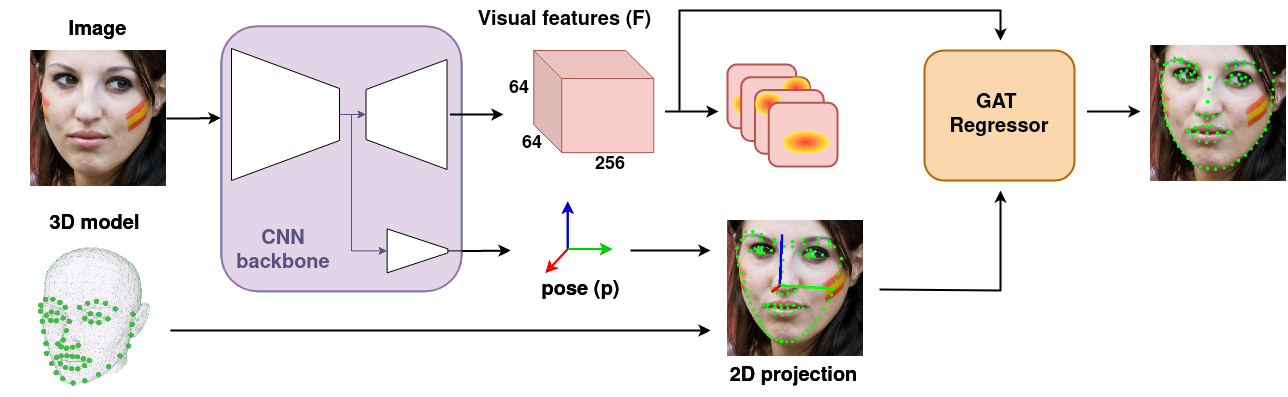}
    \caption{SPIGA workflow. Given as inputs an image and the facial 3D model, the CNN (MTN) infers the pose parameters, $\vp$, and the visual feature representation, $\mF$. Iteratively, the cascaded GAT regressor refines the initial 2D landmark projection provided by the 3D model, combining visual and structural information.}
    \label{fig:system_overview}
\end{figure}

In this section, we present a complete overview of SPIGA's implementation. Including an extended study of the CNN multi-stage backbone configuration used to provide the initialization of the 2D landmark location and the visual feature representation ($\mF$) for our GAT regressor (see Fig.~\ref{fig:system_overview}).

During training, we perform random data augmentation to the input images using the following transformations: rotation $\pm45^{\circ}$, scaling $60\pm15\%$ of the bounding box size, translation 5\% of the bounding box size, horizontal flip 50\%, blur 50\%, HSV color jittering and synthetic rectangular occlusions. Input face images are finally cropped and resized to $256\times 256$ pixels. Similarly, 64x64 output heatmaps are generated following Awing~\cite{Wang19Awing} recommendations.

\subsection{CNN Multitask Backbone}
\label{sec:cnn_backbone}
\begin{figure}[h]
    \centering
    \includegraphics[width=\columnwidth]{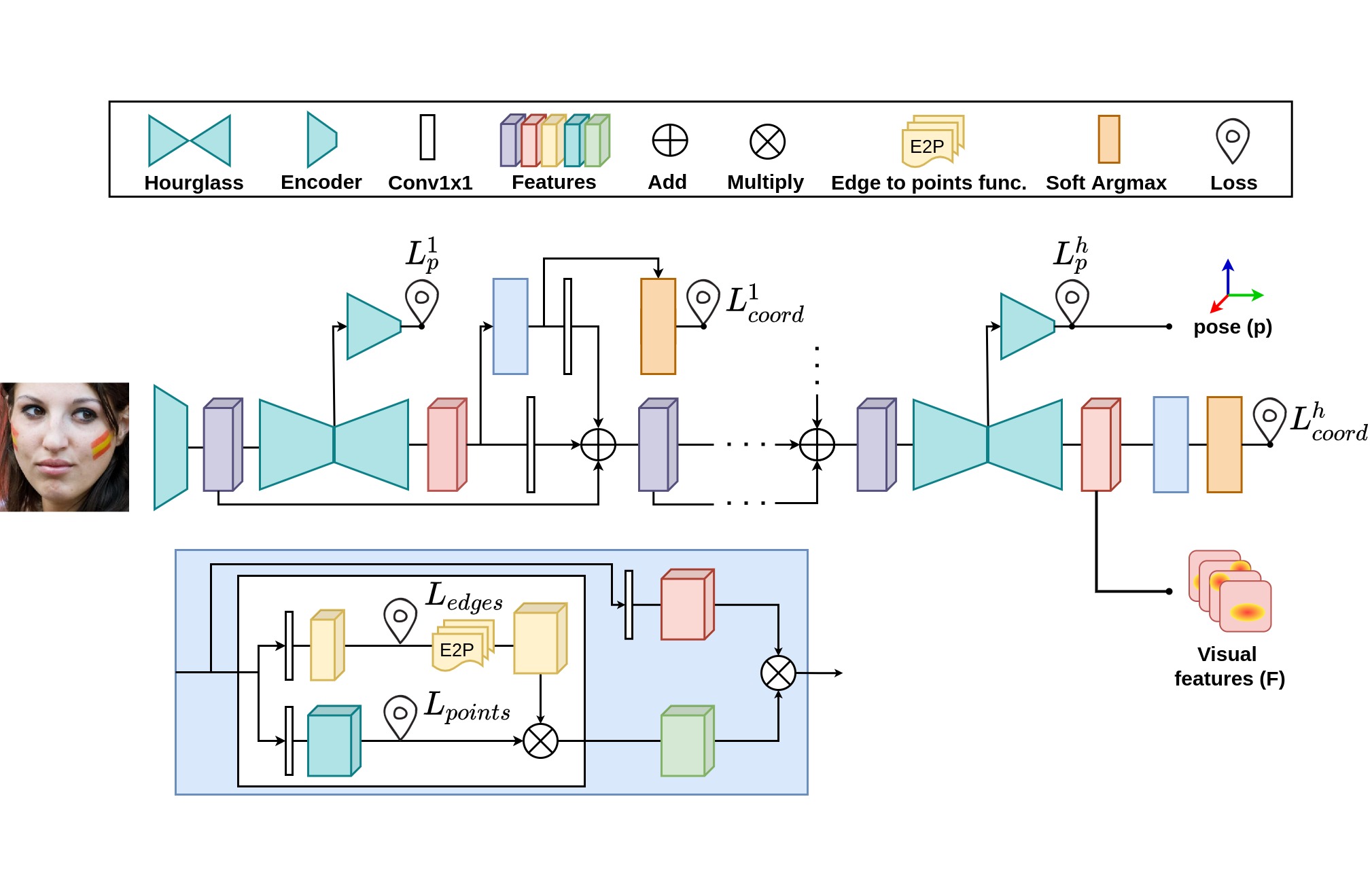}
    \caption{CNN Multitask backbone (MTL) architecture used during the fine-tuning with landmarks and pose estimation tasks.}
    \label{fig:cnn_backbone}
\end{figure}

Our backbone (MTN) consists of a cascade of $M=4$ Hourglass stages (HG) with an Attention Module, similar to the one used by~\cite{Huang21ADnet}. First, a residual encoder reduces the size of the input image from 256$\times$256 to 64$\times$64 pixels before entering the HG cascade. Each HG reduces the spatial extent of the feature maps to a resolution of 8$\times$8 at the bottleneck. Following~\cite{Valle21}, we add an encoder to each HG bottleneck to extract a 3D pose estimation head, as shown in Fig.~\ref{fig:cnn_backbone}.

We first pre-train the backbone in the landmark detection task (without the pose encoders) using the Adam optimizer during 450 epochs with an initial learning rate of $10^{-3}$ and a step decay of 0.1 at epoch 380. During training, the batch size is set to 24 and the Automatic Mixed Precision (AMP) from Pytorch is used. In Equation~\ref{eq:pretrain_lnd}, we show the loss function computed for the landmark detection task. We aggregate the losses of each HG module, represented by index $h$, doubling the loss weight of a module compared to the previous one.

\begin{equation}
\label{eq:pretrain_lnd}
    \cL_{lnd}=\sum_{h=1}^M 2^{h-1} (\lambda_{c}\cL_{coord}^h + \lambda_{att}(\cL_{points}^h  + \cL_{edges}^h)),
\end{equation}

Where $\lambda_{coord}$ and $\lambda_{att}$ are empirically set to 4 and 50, respectively. $\cL_{coord}$ is a smooth L1 function computed between the annotated and predicted landmarks coordinates. $\cL_{points}$ and $\cL_{edges}$ are Awing losses \cite{Wang19Awing} applied to the point and edges heatmaps, respectively.

Once the model has been pre-trained with landmarks, it is fine-tuned with both tasks, pose and landmarks. Sharing the same hyperparameter configuration as in the previous pre-training stage during 150 epochs, with a step decay from $10^{-3}$ to $10^{-4}$ at epoch 100. In Equation~\ref{eq:cnn_finetuning}, we show the final loss, where $\lambda_{p}$ is empirically set to 1 and $\cL_{pose}$ is the L2 loss computed for the pose estimation. Once the model is trained, we freeze the backbone to train the GAT regressor.

\begin{equation}
\label{eq:cnn_finetuning}
    \cL_{total}= \cL_{lnd} + \sum_{h=1}^M 2^{h-1} (\lambda_{p}\cL_{pose}^h)
\end{equation}

\subsection{Cascaded Regressor Based on GATs}
\label{sec:cascaded_regressor}

\begin{figure}[h]
    \centering
    \includegraphics[width=\textwidth]{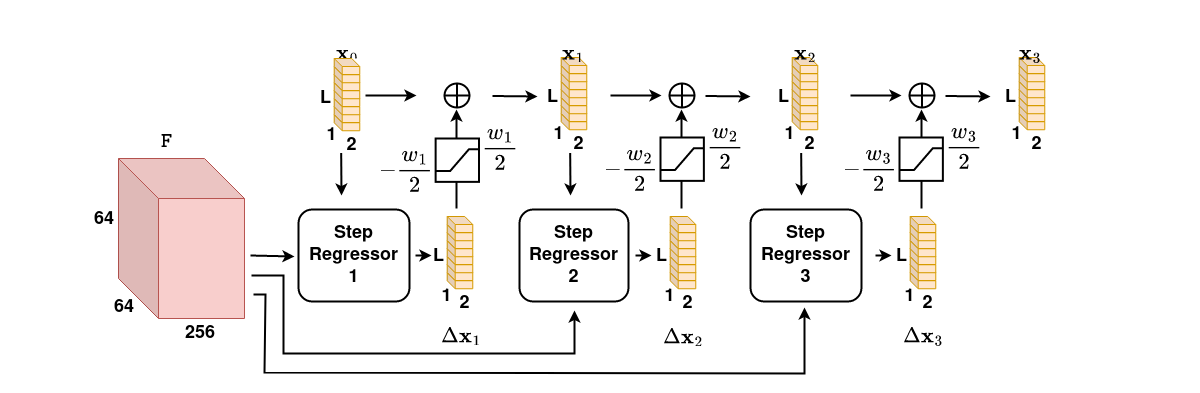}
    \caption{SPIGA cascaded regressor with the 3 steps used in the paper.}
    \label{fig:cascaded_regressor_full}
\end{figure}

\begin{figure}[h]
    \centering
    \includegraphics[width=\textwidth]{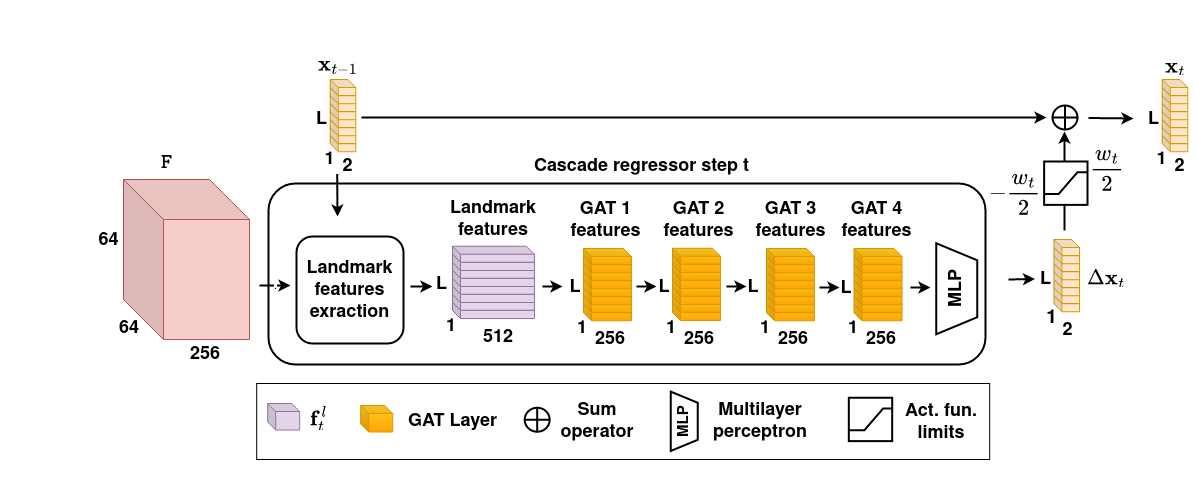}
    \caption{SPIGA step regressor with the 4 GATs layers used in the paper.}
    \label{fig:cascaded_regressor_step}
\end{figure}

The full cascaded regressor is shown in Fig.~\ref{fig:cascaded_regressor_full} and the architecture of a single-step regressor is shown in Fig.~\ref{fig:cascaded_regressor_step}. Similar to previous training configurations, the full shape regressor uses the Adam optimizer, setting an initial learning rate of $10^{-4}$ with a step decay of 0.1 at epoch 100.

\begin{figure}[h]
    \centering
    \includegraphics[width=\textwidth]{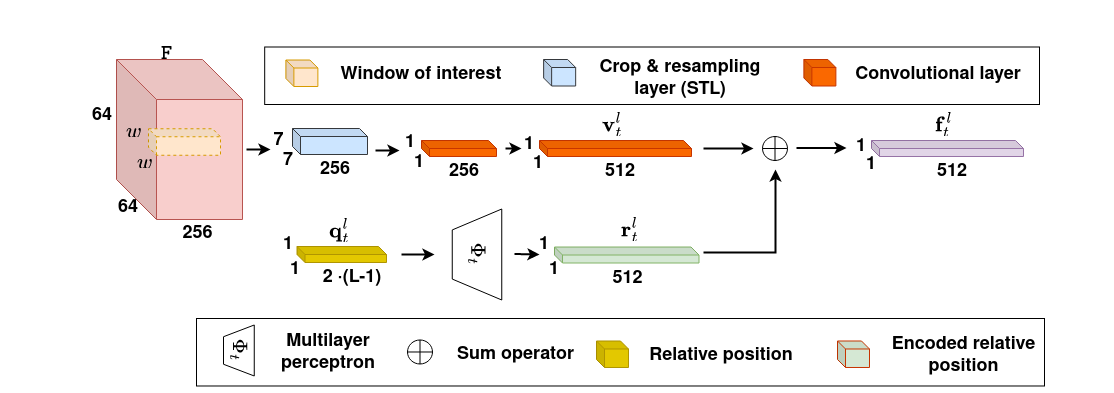}
    \caption{SPIGA extraction of visual and geometric features. Including the encoding and combination applied to get the input features of the regressor.}
    \label{fig:gnn_embeddings}
\end{figure}

The detailed extraction of visual and geometric features can be visualized in Fig.~\ref{fig:gnn_embeddings}. Including the encoding and combination applied to get the input features of the regressor. 

Let $\mF$ be the last feature map of the last stacked HG module in the MTN. We first look at a square window, $\cW_t$, of size $w_t\times w_t$ and centered at each landmark location, $\vx_{t-1}^l$ in $\mF$. We use a fixed
affine transform with the grid generator and sampler of the \emph{Spatial 
Transformer  Networks}~\cite{Jaderberg15spatial} to have a differentiable
crop operation of $\cW_t$. The crop operation re-samples $\cW_t$ to a fixed size $7\times 7 \times 256$ 
tensor, regardless of the dimension of the $w_t\times w_t$ window. 
Then, using a convolution with a $7\times 7$ kernel, a $1\times 1\times 
256$ feature map is extracted. Finally, with a $1\times 1$ convolution, we compute the 512 channels
of the visual features vector, $\vv_t^l$, corresponding to $l$-th landmark at 
step $t$. 
For each landmark $l$, we combine the visual features extracted from the backbone network, $\vv_t^l$, and the relative positional features, $\vr_t^l$, computed from $\vx_{t-1}$ (i.e. the current shape) into the encoded features, $\vf_t^l=\vv_t^l + \vr_t^l$. 

\section{Extended Experimentation}
\label{sec:results_extended}

In this section, we report an extended study of our proposal by adding new results on 
300W (public and private) and WFLW datasets. In all our tables, results ranked \first{first}, \second{second} and \third{third} are shown respectively in blue, green and red colors.
\\
\begin{table}[h]
\scriptsize
\begin{center}
\setlength\tabcolsep{4pt}
\begin{tabular}{l|c|c|c|c|c|ccc}
\hline
\multirow{2}{*}{Method} & 
\multicolumn{3}{c}{$NME_{int-ocul}$ (\%)($\downarrow$)} & 
\multicolumn{3}{|c}{$NME_{int-pupil}$ (\%)($\downarrow$)} \\
                        & Common & Challeng. & Full 
                        & Common & Challeng. & Full\\
\hline
mnv2 \cite{Fard21AMSNet}       & 3.93 & 7.52 & 4.70 & -  & - & - \\ 
SAN \cite{Dong18san}           & 3.34 & 6.60 & 3.98 &    - & -    & -    \\ 
DAN \cite{Kowalski17}          & 3.19 &  5.24 & 3.59 & 4.42 & 7.57  & 5.03 \\
TSR \cite{Lv17}                & -    & -     & -     & 4.36 & 7.56 & 4.99 \\
RAR \cite{Xiao16}              & -    & -    &  -     & 4.12 & 8.35 & 4.94 \\
LAB (4-stack) \cite{Wu18lab}   & 2.98 & 5.19 & 3.49   & 4.20 & 7.41 & 4.92 \\
FTYM \cite{Wood21fake}         & 3.09 & 4.86 & -   & - & - & -  \\
DeCaFA~\cite{Dapogny19decafa}  & 2.93 & 5.26 & 3.39 & - & - & -  \\ 
SHN \cite{Yang17}              & - & 4.90 &  - & 4.12 & 7.00 & 4.68 \\
HIHc* \cite{Lan21hih}          & 2.95 & 5.04 & 3.36 & - & - & -  \\ 
HRNetV2-W18~\cite{Wang21hrnet} & 2.87 & 5.15 & 3.32 & - &  - & - \\ 
HG$\times$2+SAAT \cite{Zhu21adversarial} & 2.87 & 5.03 & 3.29 & -    & -  & - \\ 
DCFE \cite{Valle18}            & 2.76 & 5.22 & 3.24 & 3.83 & 7.54 & 4.55 \\
AVS \cite{Qian19Avs}           & - & - &  - & 3.98 & 7.21 & 4.54 \\
PCD-CNN \cite{Kumar18a}        & - & - & - & 3.67 & 7.62 & 4.44 \\ 
SDFL \cite{LinTIP21}           & 2.88 & 4.93 & 3.28 & - & - & - \\ 
LUVLI \cite{Kumar20luvli}      & 2.76 & 5.16 & 3.23 & - & - & - \\ 
SPLT~\cite{Xia22slpt}       & 2.75 & 4.90 & 3.17 & - & - & - \\ 
3DDE \cite{Valle193dde}        & 2.69 & 4.92 & 3.13 & 3.73 & 7.10 & 4.39 \\ 
GlomFace~\cite{Zhu22glomface}  & 2.72 & 4.79 & 3.13 & - & - & - \\  
AWing \cite{Wang19Awing}       & 2.72 &       \first{4.52}   & 3.07         & 3.77         & \second{6.52}  & 4.31 \\
SLD \cite{Li20sld}             & \third{2.62} & \second{4.77}         & \third{3.04} & - &  - & - \\
DTLD-s~\cite{Li22casctransf}   & 2.67 & 4.56         & \third{3.04} & -   &  -         & - \\ 
ADNet~\cite{Huang21ADnet}      & \first{2.53} & \third{4.58} & \first{2.93} & \second{3.51} & \first{6.47}  & \second{4.08} \\
Wing \cite{Feng18wing}         & -            & -             & -            & \first{3.27} & 7.18          &  \first{4.04}  \\
\hline
\textbf{SPIGA (Ours)}                  & \second{2.59} & \textbf{4.66} & \second{2.99} & \third{3.59} &  \third{6.73} & \third{4.20} \\ 
\hline
\end{tabular}
\end{center}
\caption{Comparison against state-of-the-art on 300W public dataset.}
\label{tab:300w}
\end{table}

\textbf{300W public.} In Table~\ref{tab:300w}, we present the comparison of  state-of-the-art (SOTA) results in the 300W public. In this dataset, our approach achieves results comparable to the top performers in the literature: ADNet~\cite{Huang21ADnet} and SLD\cite{Li20sld}.
Since most images in this data set are fully visible semi-frontal faces, CNN-based methods already have a highly accurate performance (e.g. Wing). Our method is better than the other two methods using Graph Neural Networks (GraphNets), SDFL\cite{LinTIP21} and SLD\cite{Li20sld}, although results are comparable with SLD\cite{Li20sld} ($NME_{int-ocul}$ of 2.99 vs 3.04). ADNet~\cite{Huang21ADnet}, using a stacked encoder-decoder model is the SOTA and our method obtains a comparable result ($NME_{int-ocul}$ of 2.93 vs 2.99).

\textbf{300W private.} Table~\ref{tab:300w_private} shows an extended SOTA comparison in terms of \textit{NME}$_{int-ocul}$ on 300W private dataset. 

\textbf{WFLW.} In Table 5 we present an extended SOTA comparison on WLFW.

\begin{table}
\scriptsize
\begin{center}
\setlength\tabcolsep{4pt}
\begin{tabular}{l|ccc|ccc|ccc}
\hline
\multirow{3}{*}{Method} & \multicolumn{3}{c|}{Indoor} & \multicolumn{3}{c|}{Outdoor} & \multicolumn{3}{c}{Full}\\
 & $NME_{inter-ocul}$ & $AUC_8$ & $FR_8$ & $NME_{inter-ocul}$ & $AUC_8$ & $FR_8$ & $NME_{inter-ocul}$ & $AUC_8$ & $FR_8$ \\
  & $(\downarrow)$ & $(\uparrow)$ & $(\downarrow)$ & $(\downarrow)$ & $(\uparrow)$ & $(\downarrow)$ & $(\downarrow)$ & $(\uparrow)$ & $(\downarrow)$ \\
\hline
DAN~\cite{Kowalski17} & - & - & - & - & - & - & 4.30 & 47.00 & 2.67 \\
SHN~\cite{Yang17} & 4.10 & - & - & 4.00 & - & - & 4.05 & - & -  \\
DCFE~\cite{Valle18} & 3.96 & 52.28 & 2.33 & 3.81 & 52.56 & 1.33 & 3.88 & 52.42 & 1.83 \\
3DDE~\cite{Valle193dde} & 3.74 & 53.93 & 2.00& 3.71 & 53.95 & 2.66 & 3.73 & 53.94 & 2.33\\
\hline
\textbf{SPIGA (Ours)}                   & \first{3.43} & \first{57.35} & \first{1.00} & \first{3.43} & \first{57.17} & \first{0.33} & \first{3.43} & \first{57.27} & \first{0.67}\\
\hline
\end{tabular}
\end{center}
\caption{Results on 300W private test set. Face alignment methods are exclusively trained on 300W public dataset.}
\label{tab:300w_private}
\end{table}

\begin{table}
\scriptsize
\setlength{\tabcolsep}{4pt} 
\begin{center}
\begin{tabular}{c|c|ccccccc}
\hline 
Metric & Method & Testset & Pose & Expression & Illumination & Make-up & Occlusion & Blur \\
\hline
\hline 
&\multicolumn{8}{c}{Bounding boxes from WFLW benchmark} \\
\cline{2-9}
\multirow{18}{*}{$NME_{ic}$ (\%)($\downarrow$)}  
& mnv2 \cite{Fard21AMSNet}                  & 9.57 & 18.18 & 9.93 & 8.98 & 9.92 & 11.38 & 10.79\\
& LAB~\cite{Wu18lab}                        & 5.27 & 10.24 & 5.51 & 5.23 & 5.15 & 6.79 & 6.32 \\
& SAN~\cite{Dong18san}                      & 5.22 & 10.30 & 5.71 & 5.19 & 5.49 & 6.83 & 5.80 \\
& Wing~\cite{Feng18wing}                    & 5.11 & 8.75  & 5.36 & 4.93 & 5.41 & 6.37 & 5.81 \\
& 3DDE~\cite{Valle193dde}                   & 4.68 & 8.62  & 5.21 & 4.65 & 4.60 & 5.77 & 5.41 \\
& DeCaFA~\cite{Dapogny19decafa}             & 4.62 & 8.11  & 4.65 & 4.41 & 4.63 & 5.74 & 5.38 \\
& AVS+SAN~\cite{Qian19Avs}                  & 4.39 & 8.42  & 4.68 & 4.24 & 4.37 & 5.60 & 4.86 \\
& AWing~\cite{Wang19Awing}                  & 4.36 & 7.38  & 4.58 & 4.32 & 4.27 & 5.19 & 4.96 \\
\cline{2-9}
&\multicolumn{8}{c}{Bounding boxes from GT landmarks} \\
\cline{2-9}
& GlomFace~\cite{Zhu22glomface}             & 4.81 & 8.17  & - & - & - & 5.14 & - \\
& HRNetV2-W18~\cite{Wang21hrnet}            & 4.60 & 7.94 & 4.85 & 4.55 & 4.29 & 5.44 & 5.42 \\
& LUVLI~\cite{Kumar20luvli}                 & 4.37 & 7.56  & 4.77 & 4.30 & 4.33 & 5.29 & 4.94\\
& SDFL~\cite{Chu16structured}                 & 4.35 & 7.42  & 4.63 & 4.29 & 4.22 & 5.19 & 5.08\\
& AWing~\cite{Wang19Awing}              & 4.21 & \third{7.21}  & 4.46 & 4.23 & 4.02 & \third{4.99} & 4.82 \\ 
& SLD~\cite{Li20sld}                        & 4.21 & 7.36  & 4.49 & 4.12 & 4.05 & \second{4.98} & 4.82 \\
& HIHc~\cite{Lan21hih} & \third{4.18} & 7.20  & \first{4.19} & 4.45 & \second{3.97} & 5.00 & \third{4.81} \\
& ADNet~\cite{Huang21ADnet}                 & \second{4.14} & \first{6.96} & \second{4.38} & \third{4.09} & 4.05 & 5.06 & \second{4.79} \\
& DTLD-s~\cite{Li22casctransf}             & \second{4.14} & - & - & - & - & - & -\\
& SPLT~\cite{Xia22slpt}                 & \second{4.14} & \first{6.96} & \third{4.45} & \second{4.05} & \third{4.00} & 5.06 & \second{4.79}\\
\cline{2-9}
& \textbf{SPIGA (Ours)}       & \first{4.06} & \second{7.14} & \textbf{4.46} & \first{4.00} & \first{3.81} & \first{4.95} & \first{4.65}\\
\hline
\hline 
\multirow{11}{*}{$FR_{10}$ (\%)($\downarrow$)} 
& HRNetV2-W18~\cite{Wang21hrnet}            & 4.64 & 23.01 & 3.50 & 4.72 & 2.43 & 8.29 & 6.34 \\
& GlomFace~\cite{Zhu22glomface}             & 3.77 & 17.48  & - & - & - & 6.73 & - \\
& DTLD-s~\cite{Li22casctransf}             & 3.44 & - & - & - & - & - & -\\
& LUVLI~\cite{Kumar20luvli}                 & 3.12 & 15.95 & 3.18 & \third{2.15} & 3.40 & 6.39 & \third{3.23} \\
& SDFL~\cite{Chu16structured}               & 2.72 & 12.88  & \second{1.59} & 2.58 & 2.43 & 5.71 & 3.62\\
& AWing~\cite{Wang19Awing}                  & \first{2.04} & \first{9.20} & \first{1.27} & \third{2.01} & \first{0.97} & \first{4.21} & \second{2.72} \\ 
& SLD~\cite{Li20sld}                        & 3.04 & 15.95 & 2.86 & 2.72 & \second{1.46} & \third{5.29} & 4.01 \\
& HIHc~\cite{Lan21hih}                      & 2.96 & 15.03 & \second{1.59} & 2.58 & \second{1.46} & 6.11 & 3.49 \\
& ADNet~\cite{Huang21ADnet}         & \third{2.72} & 12.72 & \third{2.15} & 2.44 & \third{1.94} & 5.79 & 3.54 \\
& SPLT~\cite{Xia22slpt}             & 2.76 & \third{12.27} & 2.23 & \second{1.86} & 3.40 & 5.98 & 3.88\\
\cline{2-9}
& \textbf{SPIGA (ours)}             & \second{2.08} & \second{11.66} & \textbf{2.23} & \first{1.58} & \second{1.46} & \second{4.48} & \first{2.20}\\
\hline
\hline 
\multirow{9}{*}{$AUC_{10}$ (\%)($\uparrow$)} 
& HRNetV2-W18~\cite{Wang21hrnet}            & 52.37 & 25.06 & 51.02 & 53.26 & 54.45 & 45.85 & 45.15 \\
& LUVLI~\cite{Kumar20luvli}                 & 57.70 & 31.00 & 54.90 & 58.40 & 58.80 & 50.50 & 52.50\\
& SDFL~\cite{Chu16structured}               & 57.59 & 31.32  & 55.01 & 58.47 & 58.31 & 50.35 & 51.47\\
& AWing~\cite{Wang19Awing}                  & 58.95 & 33.37 & 57.18 & 59.58 & 60.17 & \third{52.75} & 53.93 \\ 
& SLD~\cite{Li20sld}                        & 58.93 & 31.50 & 56.63 & 59.53 & 60.38 & 52.35 & 53.29 \\
& HIHc$^1$~\cite{Lan21hih}                      & \third{59.70} & 34.20 & \first{59.00} & \second{60.60} & \third{60.40} & 52.70 & \second{54.90} \\
& ADNet~\cite{Huang21ADnet}                 & \second{60.22} & \third{34.41} & 52.34 & 58.05 & 60.07 & \second{52.95} & \third{54.80} \\
& SPLT~\cite{Xia22slpt}                 & 59.50 & \second{34.80} & \third{57.40} & \third{60.10} & \second{60.50} & 51.50 & 53.50\\
\cline{2-9}
& \textbf{SPIGA (Ours)}                             & \first{60.56} & \first{35.31} & \second{57.97} & \first{61.31} & \first{62.24} & \first{53.31} & \first{55.31}\\
\hline
\end{tabular}
\end{center}
\caption{Extended evaluation of landmark detection on WFLW.}
\label{tab:wflw}
\end{table}

\section{Extended Ablation study}

\begin{figure}
    \centering
    \includegraphics[width=\columnwidth]{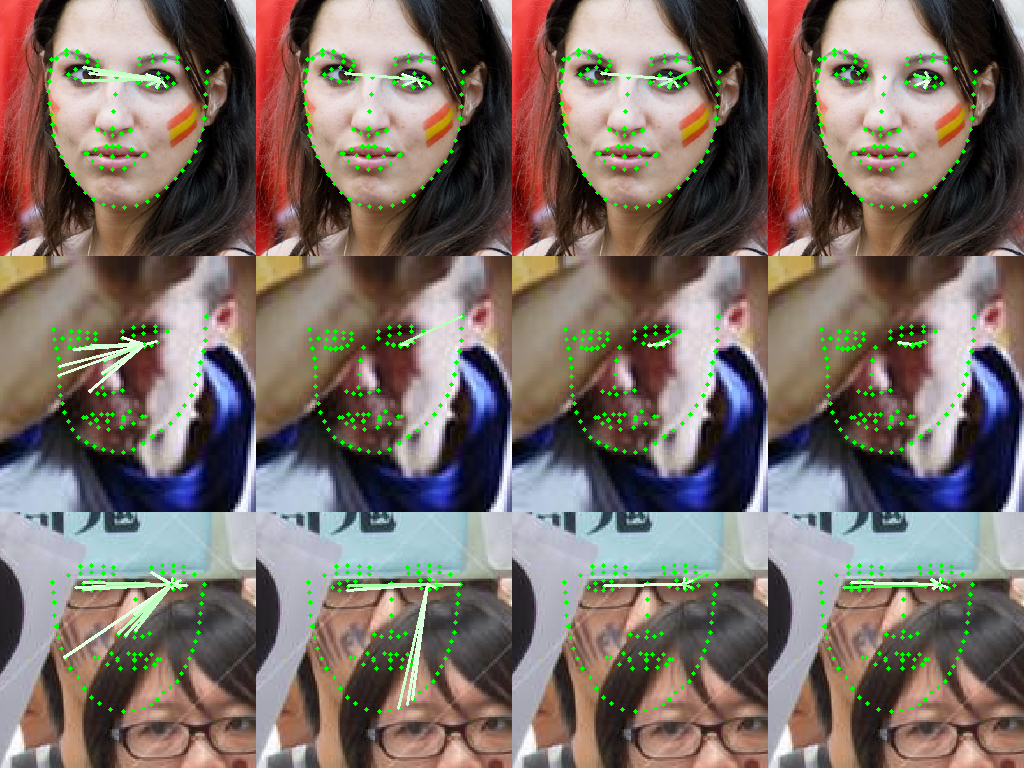}
    \caption{Attention from left eye pupil to other landmarks shown as edges. From left to right, attention at GAT layer 1, GAT layer 2, GAT layer 3 and GAT layer 4. The greener the higher is the attention. We only show edges with an attention over a threshold for clarity.}
    \label{fig:eye_attention}
\end{figure}

\begin{figure}
    \centering
    \includegraphics[width=\columnwidth]{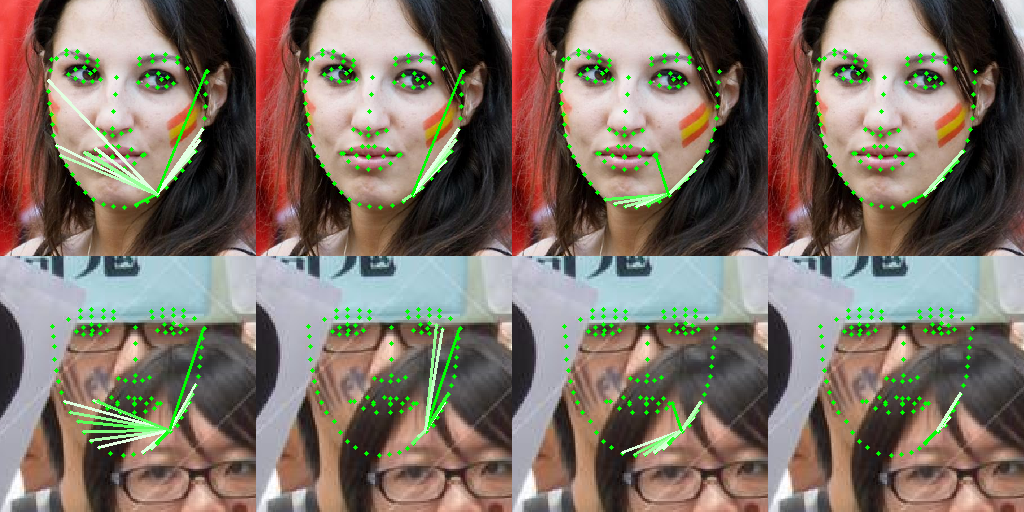}
    \caption{Attention from a landmark over the jaw to other landmarks shown as edges. From left to right, attention at GAT layer 1, GAT layer 2, GAT layer 3 and GAT layer 4. The greener the higher is the attention. We only show edges with an attention over a threshold for clarity.}
    \label{fig:cheek_attention}
\end{figure}

In this section we show more examples of the learned adjacency matrix per GAT module in the first cascade step (i.e. the attention of each landmark to others within the face graph). In Fig.~\ref{fig:eye_attention} and Fig.~\ref{fig:cheek_attention} we show the estimated landmark locations (green dots) by SPIGA. On top of landmarks locations, we show as edges the attention estimated in the first cascade regressor step for two landmarks: one from the eye pupil (see Fig.~\ref{fig:eye_attention}) and one from the jaw (see Fig.~\ref{fig:cheek_attention}). From left to right, we show the attention estimated in GAT 1 to 4. 

When we have no occlusions (see the first row in Fig.~\ref{fig:eye_attention}) to estimate the pupil features, GAT 1 looks mainly at the other eye landmarks. Then, GATs progressively pay more attention to closer landmarks and also to the other pupil. To compute the pupil displacement, GAT 4 only attends the landmarks over the same eye. Interestingly, when we have the other eye occluded (see second and third rows in Fig.~\ref{fig:eye_attention}) GAT 1 does not pay only attention to the other eye landmarks, but it looks mainly to landmarks over the nose. Finally, when we have heavy occlusions (see the last row in Fig.~\ref{fig:eye_attention}), the attention is given first to not occluded parts (i.e. nose and the other eye in GAT 1) and to landmarks over the same eye in GAT 4.

Now we study the estimated attention of a jaw landmark (see Fig.~\ref{fig:cheek_attention}). Without occlusions (first row in Fig.~\ref{fig:cheek_attention}), the jaw landmark is paying attention to the mouth and other distant jaw landmarks in GAT 1. Progressively, the attention is concentrated on closer jaw landmarks. When we have heavy occlusions, the attention is given first to non-occluded landmarks in GAT 1. This allows the first graph convolution to compute features that use non-occluded landmarks. Then, the other GATs can use closer landmarks given that the starting features were free of occlusions. 

We can conclude that the estimated attention allows us to extract occlusion-free features in the first GAT module. Then, the next GAT modules can use features from closer landmarks given the initial ones are correct.

\newpage
\section{Challenging examples}

\begin{figure}[h]
    \centering
    \subfigure{\includegraphics[width=0.175\textwidth]{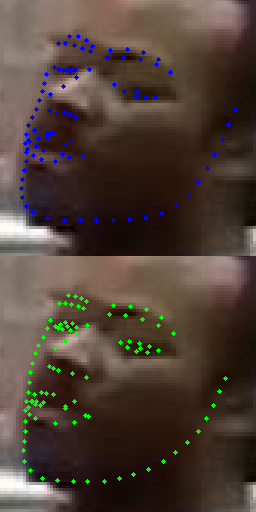}} 
    \subfigure{\includegraphics[width=0.175\textwidth]{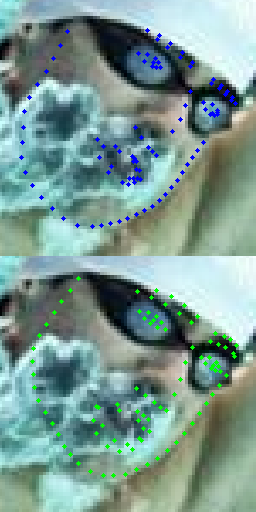}} 
    \subfigure{\includegraphics[width=0.175\textwidth]{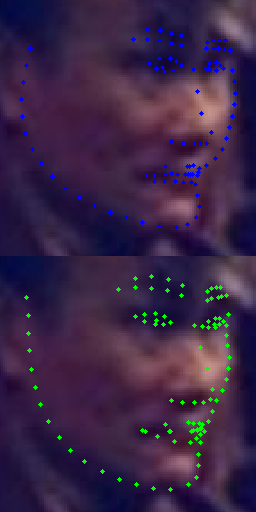}}
    \subfigure{\includegraphics[width=0.175\textwidth]{figures/figure_qualitative/expression1.png}}
    \subfigure{\includegraphics[width=0.175\textwidth]{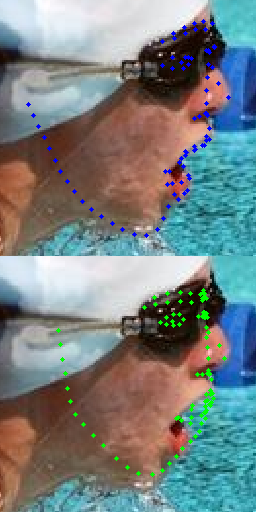}}
    \subfigure{\includegraphics[width=0.175\textwidth]{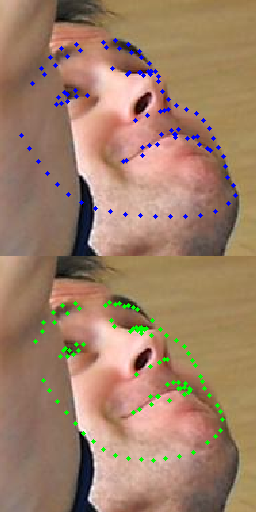}} 
    \subfigure{\includegraphics[width=0.175\textwidth]{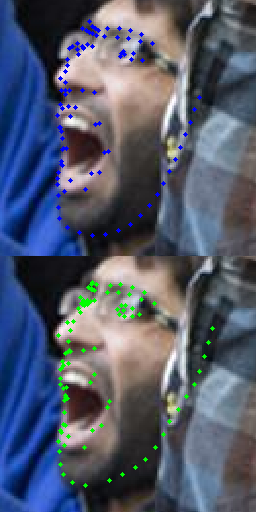}} 
    \subfigure{\includegraphics[width=0.175\textwidth]{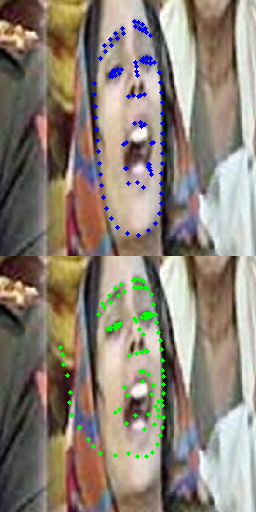}}
    \subfigure{\includegraphics[width=0.175\textwidth]{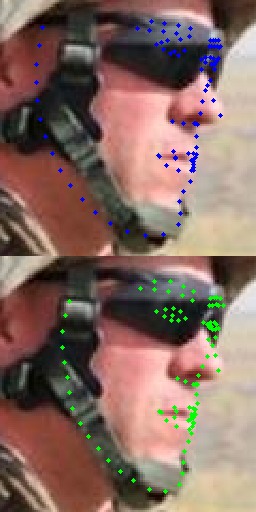}}
    \subfigure{\includegraphics[width=0.175\textwidth]{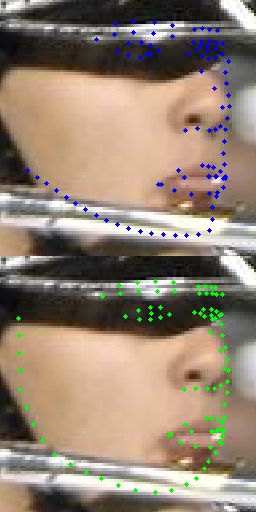}}
    \subfigure{\includegraphics[width=0.175\textwidth]{figures/figure_qualitative/occlusion9.png}} 
    \subfigure{\includegraphics[width=0.175\textwidth]{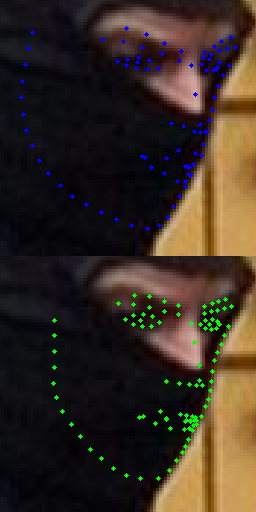}} 
    \subfigure{\includegraphics[width=0.175\textwidth]{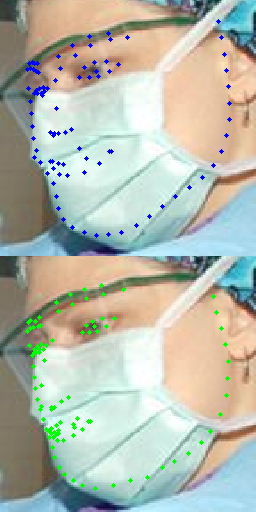}}
    \subfigure{\includegraphics[width=0.175\textwidth]{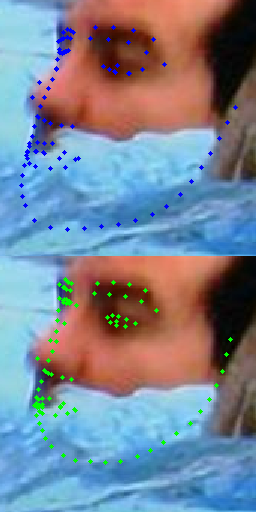}}
    \subfigure{\includegraphics[width=0.175\textwidth]{figures/figure_qualitative/occlusion7.png}}
    
    \caption{WFLW Challenging examples. In blue we show the ground truth and in green the landmark locations estimated by SPIGA.}
    \label{fig:challenging_examples}
\end{figure}

\clearpage
\bibliographystyle{splncs04}
\bibliography{faces}

%% file: labvision_defs.tex
\def\mat#1{\mathchoice{\mbox{\boldmath $\displaystyle\tt#1$}}
{\mbox{\boldmath$\textstyle\tt#1$}}
{\mbox{\boldmath$\scriptstyle\tt#1$}}
{\mbox{\boldmath$\scriptscriptstyle\tt#1$}}}

\def\vect#1{\mathchoice{\mbox{\boldmath $\displaystyle\bf#1$}}
{\mbox{\boldmath  $\textstyle\bf#1$}}
{\mbox{\boldmath  $\scriptstyle\bf#1$}}
{\mbox{\boldmath  $\scriptscriptstyle\bf#1$}}}

\def\v0{{\vect 0}}

\def\vb{{\vect b}}

\def\vf{{\vect f}}

\def\vh{{\vect h}}

\def\vm{{\vect m}}

\def\vp{{\vect p}}
\def\vq{{\vect q}}
\def\vr{{\vect r}}

\def\vv{{\vect v}}

\def\vx{{\vect x}}

\def\vX{{\vect X}}

\def\v0{{\vect 0}}

\def\mDelta{{\mat\mDelta}}

\def\mF{{\mat F}}

\def\mW{{\mat W}}

\def\m1{{\mat 1}}

\def\cA{\mathcal{A}}

\def\cL{\mathcal{L}}

\def\cW{\mathcal{W}}

\def\Reales{\mathbb{R}}


%% file: paper.bbl
\begin{thebibliography}{37}
\providecommand{\natexlab}[1]{#1}
\providecommand{\url}[1]{\texttt{#1}}
\expandafter\ifx\csname urlstyle\endcsname\relax
  \providecommand{\doi}[1]{doi: #1}\else
  \providecommand{\doi}{doi: \begingroup \urlstyle{rm}\Url}\fi

\bibitem[Burgos-Artizzu et~al.(2013)Burgos-Artizzu, Perona, and
  Dollar]{Burgos13}
Xavier~P. Burgos-Artizzu, Pietro Perona, and Piotr Dollar.
\newblock Robust face landmark estimation under occlusion.
\newblock In \emph{ICCV}, pages 1513--1520, 2013.

\bibitem[Cao et~al.(2014)Cao, Wei, Wen, and Sun]{Cao14}
Xudong Cao, Yichen Wei, Fang Wen, and Jian Sun.
\newblock Face alignment by explicit shape regression.
\newblock \emph{IJCV}, 107\penalty0 (2):\penalty0 177--190, 2014.

\bibitem[Dapogny et~al.(2019)Dapogny, Cord, and Bailly]{Dapogny19decafa}
Arnaud Dapogny, Matthieu Cord, and Kevin Bailly.
\newblock Decafa: Deep convolutional cascade for face alignment in the wild.
\newblock In \emph{ICCV}, pages 6892--6900. {IEEE}, 2019.

\bibitem[Dollar et~al.(2010)Dollar, Welinder, and Perona]{Dollar10}
Piotr Dollar, Peter Welinder, and Pietro Perona.
\newblock Cascaded pose regression.
\newblock In \emph{CVPR}, pages 1078--1085, 2010.

\bibitem[Fard et~al.(2021)Fard, Abdollahi, and Mahoor]{Fard21AMSNet}
Ali~Pourramezan Fard, Hojjat Abdollahi, and Mohammad~H. Mahoor.
\newblock Asmnet: {A} lightweight deep neural network for face alignment and
  pose estimation.
\newblock In \emph{CVPRW}, pages 1521--1530. CVF/{IEEE}, 2021.

\bibitem[Feng et~al.(2020)Feng, Kittler, Awais, and Wu]{Feng20rwing}
ZH. Feng, J.~Kittler, M.~Awais, and Xiao-Jun Wu.
\newblock Rectified wing loss for efficient and robust facial landmark
  localisation with convolutional neural networks.
\newblock \emph{IJCV}, 128:\penalty0 2126–2145, 2020.

\bibitem[Feng et~al.(2018)Feng, Kittler, Awais, Huber, and Wu]{Feng18wing}
Zhen{-}Hua Feng, Josef Kittler, Muhammad Awais, Patrik Huber, and Xiao{-}Jun
  Wu.
\newblock Wing loss for robust facial landmark localisation with convolutional
  neural networks.
\newblock In \emph{CVPR}, pages 2235--2245, 2018.

\bibitem[Honari et~al.(2016)Honari, Yosinski, Vincent, and Pal]{Honari16}
Sina Honari, Jason Yosinski, Pascal Vincent, and Christopher~J. Pal.
\newblock Recombinator networks: Learning coarse-to-fine feature aggregation.
\newblock In \emph{CVPR}, pages 5743--5752, 2016.

\bibitem[Huang et~al.(2020)Huang, Deng, Shen, Zhang, and Ye]{Huang20propnet}
Xiehe Huang, Weihong Deng, Haifeng Shen, Xiubao Zhang, and Jieping Ye.
\newblock Propagationnet: Propagate points to curve to learn structure
  information.
\newblock In \emph{CVPR}, June 2020.

\bibitem[Huang et~al.(2021)Huang, Yang, Li, Kim, and Wei]{Huang21ADnet}
Yangyu Huang, Hao Yang, Chong Li, Jongyoo Kim, and Fangyun Wei.
\newblock Adnet: Leveraging error-bias towards normal direction in face
  alignment.
\newblock In \emph{ICCV}, pages 3080--3090, October 2021.

\bibitem[Jaderberg et~al.(2015)Jaderberg, Simonyan, Zisserman, and
  Kavukcuoglu]{Jaderberg15spatial}
Max Jaderberg, Karen Simonyan, Andrew Zisserman, and Koray Kavukcuoglu.
\newblock Spatial transformer networks.
\newblock In Corinna Cortes, Neil~D. Lawrence, Daniel~D. Lee, Masashi Sugiyama,
  and Roman Garnett, editors, \emph{NeurIPS}, pages 2017--2025, 2015.

\bibitem[Kowalski et~al.(2017)Kowalski, Naruniec, and Trzcinski]{Kowalski17}
Marek Kowalski, Jacek Naruniec, and Tomasz Trzcinski.
\newblock Deep alignment network: {A} convolutional neural network for robust
  face alignment.
\newblock In \emph{CVPRW}, pages 2034--2043, 2017.

\bibitem[Kumar et~al.(2020)Kumar, Marks, Mou, Wang, Jones, Cherian,
  Koike-Akino, Liu, and Feng]{Kumar20luvli}
Abhinav Kumar, Tim~K. Marks, Wenxuan Mou, Ye~Wang, Michael Jones, Anoop
  Cherian, Toshiaki Koike-Akino, Xiaoming Liu, and Chen Feng.
\newblock Luvli face alignment: Estimating landmarks’ location, uncertainty,
  and visibility likelihood.
\newblock In \emph{CVPR}, pages 8233--8243, 2020.

\bibitem[Lan et~al.(2021)Lan, Hu, and Cheng]{Lan21hih}
Xing Lan, Qinghao Hu, and Jian Cheng.
\newblock Revisting quantization error in face alignment.
\newblock In \emph{ICCVW}, pages 1521--1530, October 2021.

\bibitem[Li et~al.(2022)Li, Guo, Rhee, Han, and Han]{Li22casctransf}
Hui Li, Zidong Guo, Seon-Min Rhee, Seungju Han, and Jae-Joon Han.
\newblock Towards accurate facial landmark detection via cascaded transformers.
\newblock In \emph{Proceedings of the IEEE/CVF CVPR}, pages 4176--4185, June
  2022.

\bibitem[Li et~al.(2020)Li, Lu, Zheng, Liao, Lin, Luo, Cheng, Xiao, Lu, Kuo,
  and Miao]{Li20sld}
Weijian Li, Yuhang Lu, Kang Zheng, Haofu Liao, Chihung Lin, Jiebo Luo, Chi-Tung
  Cheng, Jing Xiao, Le~Lu, Chang-Fu Kuo, and Shun Miao.
\newblock Structured landmark detection via topology-adapting deep graph
  learning.
\newblock In Andrea Vedaldi, Horst Bischof, Thomas Brox, and Jan-Michael Frahm,
  editors, \emph{ECCV}, pages 266--283. Springer International Publishing,
  2020.

\bibitem[Lin et~al.(2021)Lin, Zhu, Wang, Liao, Qian, Lu, and Zhou]{LinTIP21}
Chunze Lin, Beier Zhu, Quan Wang, Renjie Liao, Chen Qian, Jiwen Lu, and Jie
  Zhou.
\newblock Structure-coherent deep feature learning for robust face alignment.
\newblock \emph{IEEE TIP}, 30:\penalty0 5313--5326, 2021.

\bibitem[Moskvyak et~al.(2021)Moskvyak, Maire, Dayoub, and
  Baktashmotlagh]{MoskvyakWACV21}
Olga Moskvyak, Frederic Maire, Feras Dayoub, and Mahsa Baktashmotlagh.
\newblock Keypoint-aligned embeddings for image retrieval and
  re-identification.
\newblock In \emph{WACV}, pages 676--685, January 2021.

\bibitem[Qian et~al.(2019)Qian, Sun, Wu, Qian, and Jia]{Qian19Avs}
Shengju Qian, Keqiang Sun, Wayne Wu, Chen Qian, and Jiaya Jia.
\newblock Aggregation via separation: Boosting facial landmark detector with
  semi-supervised style translation.
\newblock In \emph{ICCV}, October 2019.

\bibitem[Sagonas et~al.(2016)Sagonas, Tzimiropoulos, Zafeiriou, and
  Pantic]{Sagonas16}
Christos Sagonas, Georgios Tzimiropoulos, Stefanos Zafeiriou, and Maja Pantic.
\newblock 300 faces in-the-wild challenge: database and results.
\newblock \emph{IVC}, 47:\penalty0 3--18, 2016.

\bibitem[Santoro et~al.(2017)Santoro, Raposo, Barrett, Malinowski, Pascanu,
  Battaglia, and Lillicrap]{Santoro17}
A.~Santoro, D.~Raposo, D.~G Barrett, M.~Malinowski, R.~Pascanu, P.~Battaglia,
  and T.~Lillicrap.
\newblock A simple neural network module for relational reasoning.
\newblock In \emph{NeurIPS}, 2017.

\bibitem[Sarlin et~al.(2020)Sarlin, DeTone, Malisiewicz, and
  Rabinovich]{Sarlin20superglue}
Paul-Edouard Sarlin, Daniel DeTone, Tomasz Malisiewicz, and Andrew Rabinovich.
\newblock Superglue: Learning feature matching with graph neural networks.
\newblock In \emph{CVPR}, June 2020.

\bibitem[Sun et~al.(2019)Sun, Li, Huan, Liu, and Han]{Sun19}
Ning Sun, Qi~Li, Ruizhi Huan, Jixin Liu, and Guang Han.
\newblock Deep spatial-temporal feature fusion for facial expression
  recognition in static images.
\newblock \emph{PRL}, 119:\penalty0 49--61, 2019.

\bibitem[Trigeorgis et~al.(2016)Trigeorgis, Snape, Nicolaou, Antonakos, and
  Zafeiriou]{Trigeorgis16}
George Trigeorgis, Patrick Snape, Mihalis~A. Nicolaou, Epameinondas Antonakos,
  and Stefanos Zafeiriou.
\newblock Mnemonic descent method: {A} recurrent process applied for end-to-end
  face alignment.
\newblock In \emph{CVPR}, pages 4177--4187, 2016.

\bibitem[Valle et~al.(2018)Valle, Buenaposada, Vald{\'{e}}s, and
  Baumela]{Valle18}
Roberto Valle, Jos{\'{e}}~M. Buenaposada, Antonio Vald{\'{e}}s, and Luis
  Baumela.
\newblock A deeply-initialized coarse-to-fine ensemble of regression trees for
  face alignment.
\newblock In \emph{ECCV}, pages 609--624, 2018.

\bibitem[Valle et~al.(2019)Valle, Buenaposada, Vald{\'{e}}s, and
  Baumela]{Valle193dde}
Roberto Valle, Jos{\'{e}}~M. Buenaposada, Antonio Vald{\'{e}}s, and Luis
  Baumela.
\newblock Face alignment using a {3D} deeply-initialized ensemble of regression
  trees.
\newblock \emph{CVIU}, 189:\penalty0 102846, 2019.

\bibitem[Valle et~al.(2021)Valle, Buenaposada, and Baumela]{Valle21}
Roberto Valle, Jos{\'{e}}~M. Buenaposada, and Luis Baumela.
\newblock Multi-task head pose estimation in-the-wild.
\newblock \emph{IEEE TPAMI}, 43\penalty0 (8):\penalty0 2874--2881, 2021.

\bibitem[Velickovic et~al.(2018)Velickovic, Cucurull, Casanova, Romero, Lio,
  and Bengio]{Velickovic18gats}
Petar Velickovic, Guillem Cucurull, Arantxa Casanova, Adriana Romero, Pietro
  Lio, and Yoshua Bengio.
\newblock Graph attention networks.
\newblock In \emph{ICLR}, 2018.

\bibitem[Wang et~al.(2021)Wang, Sun, Cheng, Jiang, Deng, Zhao, Liu, Mu, Tan,
  Wang, Liu, and Xiao]{Wang21hrnet}
Jingdong Wang, Ke~Sun, Tianheng Cheng, Borui Jiang, Chaorui Deng, Yang Zhao,
  Dong Liu, Yadong Mu, Mingkui Tan, Xinggang Wang, Wenyu Liu, and Bin Xiao.
\newblock Deep high-resolution representation learning for visual recognition.
\newblock \emph{IEEE TPAMI}, 43\penalty0 (10):\penalty0 3349--3364, 2021.

\bibitem[Wang et~al.(2019)Wang, Bo, and Fuxin]{Wang19Awing}
Xinyao Wang, Liefeng Bo, and Li~Fuxin.
\newblock Adaptive wing loss for robust face alignment via heatmap regression.
\newblock In \emph{ICCV}, October 2019.

\bibitem[Wu et~al.(2018)Wu, Qian, Yang, Wang, Cai, and Zhou]{Wu18lab}
Wayne Wu, Chen Qian, Shuo Yang, Quan Wang, Yici Cai, and Qiang Zhou.
\newblock Look at boundary: {A} boundary-aware face alignment algorithm.
\newblock In \emph{CVPR}, pages 2129--2138, 2018.

\bibitem[Xia et~al.(2022)Xia, Qu, Huang, Zhang, Wang, and Xu]{Xia22slpt}
Jiahao Xia, Weiwei Qu, Wenjian Huang, Jianguo Zhang, Xi~Wang, and Min Xu.
\newblock Sparse local patch transformer for robust face alignment and
  landmarks inherent relation learning.
\newblock In \emph{Proceedings of the IEEE/CVF CVPR}, pages 4052--4061, June
  2022.

\bibitem[Xu and Kakadiaris(2017)]{Xu17JFA}
Xiang Xu and Ioannis~A. Kakadiaris.
\newblock Joint head pose estimation and face alignment framework using global
  and local {CNN} features.
\newblock In \emph{IEEE Int. Conf. on Automatic Face and Gesture Recognition},
  pages 642--649. {IEEE} Computer Society, 2017.

\bibitem[Yang et~al.(2015)Yang, Mou, Zhang, Patras, Gunes, and
  Robinson]{Yang15}
Heng Yang, Wenxuan Mou, Yichi Zhang, Ioannis Patras, Hatice Gunes, and Peter
  Robinson.
\newblock Face alignment assisted by head pose estimation.
\newblock In \emph{BMVC}, pages 130.1--130.13, 2015.

\bibitem[Zhang et~al.(2020)Zhang, Zeng, Wang, Pan, Liu, Liu, Ding, and
  Fan]{Zhang20Freenet}
Jiangning Zhang, Xianfang Zeng, Mengmeng Wang, Yusu Pan, Liang Liu, Yong Liu,
  Yu~Ding, and Changjie Fan.
\newblock Freenet: Multi-identity face reenactment.
\newblock In \emph{CVPR}, pages 5325--5334, 2020.

\bibitem[Zhu et~al.(2021)Zhu, Li, Li, and Dai]{Zhu21adversarial}
Congcong Zhu, Xiaoqiang Li, Jide Li, and Songmin Dai.
\newblock Improving robustness of facial landmark detection by defending
  against adversarial attacks.
\newblock In \emph{ICCV}, pages 11751--11760, October 2021.

\bibitem[Zhu et~al.(2022)Zhu, Wan, Xie, Li, and Gu]{Zhu22glomface}
Congcong Zhu, Xintong Wan, Shaorong Xie, Xiaoqiang Li, and Yinzheng Gu.
\newblock Occlusion-robust face alignment using a viewpoint-invariant
  hierarchical network architecture.
\newblock In \emph{Proceedings of the IEEE/CVF CVPR}, pages 11112--11121, June
  2022.

\end{thebibliography}


\begin{thebibliography}{28}
\providecommand{\natexlab}[1]{#1}
\providecommand{\url}[1]{\texttt{#1}}
\expandafter\ifx\csname urlstyle\endcsname\relax
  \providecommand{\doi}[1]{doi: #1}\else
  \providecommand{\doi}{doi: \begingroup \urlstyle{rm}\Url}\fi

\bibitem[Chu et~al.(2016)Chu, Ouyang, Li, and Wang]{Chu16structured}
Xiao Chu, Wanli Ouyang, Hongsheng Li, and Xiaogang Wang.
\newblock Structured feature learning for pose estimation.
\newblock In \emph{CVPR}, June 2016.

\bibitem[Dapogny et~al.(2019)Dapogny, Cord, and Bailly]{Dapogny19decafa}
Arnaud Dapogny, Matthieu Cord, and Kevin Bailly.
\newblock Decafa: Deep convolutional cascade for face alignment in the wild.
\newblock In \emph{ICCV}, pages 6892--6900. {IEEE}, 2019.

\bibitem[Dong et~al.(2018)Dong, Yan, Ouyang, and Yang]{Dong18san}
Xuanyi Dong, Yan Yan, Wanli Ouyang, and Yi~Yang.
\newblock Style aggregated network for facial landmark detection.
\newblock In \emph{CVPR}, pages 379--388, 2018.

\bibitem[Fard et~al.(2021)Fard, Abdollahi, and Mahoor]{Fard21AMSNet}
Ali~Pourramezan Fard, Hojjat Abdollahi, and Mohammad~H. Mahoor.
\newblock Asmnet: {A} lightweight deep neural network for face alignment and
  pose estimation.
\newblock In \emph{CVPRW}, pages 1521--1530. CVF/{IEEE}, 2021.

\bibitem[Feng et~al.(2018)Feng, Kittler, Awais, Huber, and Wu]{Feng18wing}
Zhen{-}Hua Feng, Josef Kittler, Muhammad Awais, Patrik Huber, and Xiao{-}Jun
  Wu.
\newblock Wing loss for robust facial landmark localisation with convolutional
  neural networks.
\newblock In \emph{CVPR}, pages 2235--2245, 2018.

\bibitem[Huang et~al.(2021)Huang, Yang, Li, Kim, and Wei]{Huang21ADnet}
Yangyu Huang, Hao Yang, Chong Li, Jongyoo Kim, and Fangyun Wei.
\newblock Adnet: Leveraging error-bias towards normal direction in face
  alignment.
\newblock In \emph{ICCV}, pages 3080--3090, October 2021.

\bibitem[Jaderberg et~al.(2015)Jaderberg, Simonyan, Zisserman, and
  Kavukcuoglu]{Jaderberg15spatial}
Max Jaderberg, Karen Simonyan, Andrew Zisserman, and Koray Kavukcuoglu.
\newblock Spatial transformer networks.
\newblock In Corinna Cortes, Neil~D. Lawrence, Daniel~D. Lee, Masashi Sugiyama,
  and Roman Garnett, editors, \emph{NeurIPS}, pages 2017--2025, 2015.

\bibitem[Kowalski et~al.(2017)Kowalski, Naruniec, and Trzcinski]{Kowalski17}
Marek Kowalski, Jacek Naruniec, and Tomasz Trzcinski.
\newblock Deep alignment network: {A} convolutional neural network for robust
  face alignment.
\newblock In \emph{CVPRW}, pages 2034--2043, 2017.

\bibitem[Kumar et~al.(2020)Kumar, Marks, Mou, Wang, Jones, Cherian,
  Koike-Akino, Liu, and Feng]{Kumar20luvli}
Abhinav Kumar, Tim~K. Marks, Wenxuan Mou, Ye~Wang, Michael Jones, Anoop
  Cherian, Toshiaki Koike-Akino, Xiaoming Liu, and Chen Feng.
\newblock Luvli face alignment: Estimating landmarks’ location, uncertainty,
  and visibility likelihood.
\newblock In \emph{CVPR}, pages 8233--8243, 2020.

\bibitem[Kumar and Chellappa(2018)]{Kumar18a}
Amit Kumar and Rama Chellappa.
\newblock Disentangling {3D} pose in a dendritic {CNN} for unconstrained {2D}
  face alignment.
\newblock In \emph{CVPR}, pages 430--439, 2018.

\bibitem[Lan et~al.(2021)Lan, Hu, and Cheng]{Lan21hih}
Xing Lan, Qinghao Hu, and Jian Cheng.
\newblock Revisting quantization error in face alignment.
\newblock In \emph{ICCVW}, pages 1521--1530, October 2021.

\bibitem[Li et~al.(2022)Li, Guo, Rhee, Han, and Han]{Li22casctransf}
Hui Li, Zidong Guo, Seon-Min Rhee, Seungju Han, and Jae-Joon Han.
\newblock Towards accurate facial landmark detection via cascaded transformers.
\newblock In \emph{Proceedings of the IEEE/CVF CVPR}, pages 4176--4185, June
  2022.

\bibitem[Li et~al.(2020)Li, Lu, Zheng, Liao, Lin, Luo, Cheng, Xiao, Lu, Kuo,
  and Miao]{Li20sld}
Weijian Li, Yuhang Lu, Kang Zheng, Haofu Liao, Chihung Lin, Jiebo Luo, Chi-Tung
  Cheng, Jing Xiao, Le~Lu, Chang-Fu Kuo, and Shun Miao.
\newblock Structured landmark detection via topology-adapting deep graph
  learning.
\newblock In Andrea Vedaldi, Horst Bischof, Thomas Brox, and Jan-Michael Frahm,
  editors, \emph{ECCV}, pages 266--283. Springer International Publishing,
  2020.

\bibitem[Lin et~al.(2021)Lin, Zhu, Wang, Liao, Qian, Lu, and Zhou]{LinTIP21}
Chunze Lin, Beier Zhu, Quan Wang, Renjie Liao, Chen Qian, Jiwen Lu, and Jie
  Zhou.
\newblock Structure-coherent deep feature learning for robust face alignment.
\newblock \emph{IEEE TIP}, 30:\penalty0 5313--5326, 2021.

\bibitem[Lv et~al.(2017)Lv, Shao, Xing, Cheng, and Zhou]{Lv17}
Jiangjing Lv, Xiaohu Shao, Junliang Xing, Cheng Cheng, and Xi~Zhou.
\newblock A deep regression architecture with two-stage re-initialization for
  high performance facial landmark detection.
\newblock In \emph{CVPR}, pages 3691--3700, 2017.

\bibitem[Qian et~al.(2019)Qian, Sun, Wu, Qian, and Jia]{Qian19Avs}
Shengju Qian, Keqiang Sun, Wayne Wu, Chen Qian, and Jiaya Jia.
\newblock Aggregation via separation: Boosting facial landmark detector with
  semi-supervised style translation.
\newblock In \emph{ICCV}, October 2019.

\bibitem[Valle et~al.(2018)Valle, Buenaposada, Vald{\'{e}}s, and
  Baumela]{Valle18}
Roberto Valle, Jos{\'{e}}~M. Buenaposada, Antonio Vald{\'{e}}s, and Luis
  Baumela.
\newblock A deeply-initialized coarse-to-fine ensemble of regression trees for
  face alignment.
\newblock In \emph{ECCV}, pages 609--624, 2018.

\bibitem[Valle et~al.(2019)Valle, Buenaposada, Vald{\'{e}}s, and
  Baumela]{Valle193dde}
Roberto Valle, Jos{\'{e}}~M. Buenaposada, Antonio Vald{\'{e}}s, and Luis
  Baumela.
\newblock Face alignment using a {3D} deeply-initialized ensemble of regression
  trees.
\newblock \emph{CVIU}, 189:\penalty0 102846, 2019.

\bibitem[Valle et~al.(2021)Valle, Buenaposada, and Baumela]{Valle21}
Roberto Valle, Jos{\'{e}}~M. Buenaposada, and Luis Baumela.
\newblock Multi-task head pose estimation in-the-wild.
\newblock \emph{IEEE TPAMI}, 43\penalty0 (8):\penalty0 2874--2881, 2021.

\bibitem[Wang et~al.(2021)Wang, Sun, Cheng, Jiang, Deng, Zhao, Liu, Mu, Tan,
  Wang, Liu, and Xiao]{Wang21hrnet}
Jingdong Wang, Ke~Sun, Tianheng Cheng, Borui Jiang, Chaorui Deng, Yang Zhao,
  Dong Liu, Yadong Mu, Mingkui Tan, Xinggang Wang, Wenyu Liu, and Bin Xiao.
\newblock Deep high-resolution representation learning for visual recognition.
\newblock \emph{IEEE TPAMI}, 43\penalty0 (10):\penalty0 3349--3364, 2021.

\bibitem[Wang et~al.(2019)Wang, Bo, and Fuxin]{Wang19Awing}
Xinyao Wang, Liefeng Bo, and Li~Fuxin.
\newblock Adaptive wing loss for robust face alignment via heatmap regression.
\newblock In \emph{ICCV}, October 2019.

\bibitem[Wood et~al.(2021)Wood, Baltrusaitis, Hewitt, Dziadzio, Johnson,
  Estellers, Cashman, and Shotton]{Wood21fake}
Erroll Wood, Tadas Baltrusaitis, Charlie Hewitt, Sebastian Dziadzio, Matthew
  Johnson, Virginia Estellers, Tom Cashman, and Jamie Shotton.
\newblock Fake it till you make it: Face analysis in the wild using synthetic
  data alone.
\newblock In \emph{ICCV}, October 2021.

\bibitem[Wu et~al.(2018)Wu, Qian, Yang, Wang, Cai, and Zhou]{Wu18lab}
Wayne Wu, Chen Qian, Shuo Yang, Quan Wang, Yici Cai, and Qiang Zhou.
\newblock Look at boundary: {A} boundary-aware face alignment algorithm.
\newblock In \emph{CVPR}, pages 2129--2138, 2018.

\bibitem[Xia et~al.(2022)Xia, Qu, Huang, Zhang, Wang, and Xu]{Xia22slpt}
Jiahao Xia, Weiwei Qu, Wenjian Huang, Jianguo Zhang, Xi~Wang, and Min Xu.
\newblock Sparse local patch transformer for robust face alignment and
  landmarks inherent relation learning.
\newblock In \emph{Proceedings of the IEEE/CVF CVPR}, pages 4052--4061, June
  2022.

\bibitem[Xiao et~al.(2016)Xiao, Feng, Xing, Lai, Yan, and Kassim]{Xiao16}
Shengtao Xiao, Jiashi Feng, Junliang Xing, Hanjiang Lai, Shuicheng Yan, and
  Ashraf~A. Kassim.
\newblock Robust facial landmark detection via recurrent attentive-refinement
  networks.
\newblock In \emph{ECCV}, pages 57--72, 2016.

\bibitem[Yang et~al.(2017)Yang, Liu, and Zhang]{Yang17}
Jing Yang, Qingshan Liu, and Kaihua Zhang.
\newblock Stacked hourglass network for robust facial landmark localisation.
\newblock In \emph{CVPRW}, pages 2025--2033, 2017.

\bibitem[Zhu et~al.(2021)Zhu, Li, Li, and Dai]{Zhu21adversarial}
Congcong Zhu, Xiaoqiang Li, Jide Li, and Songmin Dai.
\newblock Improving robustness of facial landmark detection by defending
  against adversarial attacks.
\newblock In \emph{ICCV}, pages 11751--11760, October 2021.

\bibitem[Zhu et~al.(2022)Zhu, Wan, Xie, Li, and Gu]{Zhu22glomface}
Congcong Zhu, Xintong Wan, Shaorong Xie, Xiaoqiang Li, and Yinzheng Gu.
\newblock Occlusion-robust face alignment using a viewpoint-invariant
  hierarchical network architecture.
\newblock In \emph{Proceedings of the IEEE/CVF CVPR}, pages 11112--11121, June
  2022.

\end{thebibliography}
